\documentclass{article} 

\usepackage{iclr2024_conference,times}
\usepackage{graphicx}
\usepackage{hyperref}
\usepackage{url}
\usepackage{subcaption}
\usepackage{wrapfig}
\usepackage{bbm}
\usepackage{mathtools}
\usepackage[T1]{fontenc}
\usepackage[export]{adjustbox}

\newcommand{\partitle}[1]{\textbf{#1}}

\iclrfinalcopy

\usepackage{amsmath,amsfonts,bm}

\def\eqref#1{equation~\ref{#1}}

\def\1{\bm{1}}

\DeclareMathAlphabet{\mathsfit}{\encodingdefault}{\sfdefault}{m}{sl}
\SetMathAlphabet{\mathsfit}{bold}{\encodingdefault}{\sfdefault}{bx}{n}

\DeclareMathOperator*{\argmax}{arg\,max}

\title{Reward Design for Justifiable Sequential Decision-Making}

\author{
    Aleksa Sukovic \textsuperscript{\rm 1, 2} \; Goran Radanovic \textsuperscript{\rm 1} \\
    \;Max Planck Institute for Software Systems \textsuperscript{\rm 1} \; Saarland University \textsuperscript{\rm 2} \\
    \texttt{\{asukovic, gradanovic\}@mpi-sws.org}
}

\begin{document}

\maketitle

\begin{abstract}
Equipping agents with the capacity to justify made decisions using supporting evidence represents a cornerstone of accountable decision-making. Furthermore, ensuring that justifications are in line with human expectations and societal norms is vital, especially in high-stakes situations such as healthcare. In this work, we propose the use of a debate-based reward model for reinforcement learning agents, where the outcome of a zero-sum debate game quantifies the justifiability of a decision in a particular state. This reward model is then used to train a justifiable policy, whose decisions can be more easily corroborated with supporting evidence. In the debate game, two argumentative agents take turns providing supporting evidence for two competing decisions. Given the proposed evidence, a proxy of a human judge evaluates which decision is better justified. We demonstrate the potential of our approach in learning policies for prescribing and justifying treatment decisions of septic patients. We show that augmenting the reward with the feedback signal generated by the debate-based reward model yields policies highly favored by the judge when compared to the policy obtained solely from the environment rewards, while hardly sacrificing any performance. Moreover, in terms of the overall performance and justifiability of trained policies, the debate-based feedback is comparable to the feedback obtained from an ideal judge proxy that evaluates decisions using the full information encoded in the state. This suggests that the debate game outputs key information contained in states that is most relevant for evaluating decisions, which in turn substantiates the practicality of combining our approach with human-in-the-loop evaluations. Lastly, we showcase that agents trained via multi-agent debate learn to propose evidence that is resilient to refutations and closely aligns with human preferences.
\end{abstract}

\section{Introduction}
Reinforcement learning (RL) has been achieving impressive successes in a wide range of complex domains. However, specifying a reward function which incentivizes RL agents to exhibit a desired behavior remains difficult \citep{leike1811scalable}. Prior work proposes several approaches that address these difficulties \citep{kwon2023reward,bahdanau2018learning,jothimurugan2019composable}, including those based on learning from pairwise preferences \citep{christiano2017deep}. However, such reward models are not informative enough for agents to learn how to substantiate their decisions with supporting evidence -- a key property needed for accountable decision-making \citep{bovens2007analysing}. Hence, we ask the following question:  \textit{How can we design rewards that incentivize agents to carry out a given task, but through decisions that can be justified?} 

To answer this questions, we consider a setting in which an RL agent acts as a principal, influencing the state of the world, while a human agent acts as a verifier responsible for validating the justifiability of the decisions taken by the RL agent, based on the evidence provided. This scenario mirrors common real-world situations where accountability is critical, including healthcare scenarios where doctors are tasked with scrutinizing the validity of automated decisions. We recognize three important properties that the provided evidence should satisfy. First, it should reflect the underlying state of the world, i.e., include the key information based on which an action was taken. A naive solution is to provide the full state as evidence. However, this discards the fact that the human, as a suboptimal-decision maker, may not easily reason about the taken decision because the state might be too large or otherwise incomprehensible. This brings us to the second property: the provided evidence should also be concise and only reflect the most relevant information. The third property builds on the second one: given that the provided evidence contains only partial information about the state, this information should not be easy to refute. More specifically, additional information about the underlying state, e.g., additional evidence, should not change the human's judgment. Therefore, the overall challenge is to design a framework which can enable such justifications through a reward model that incentivizes both performance and justifiability.  

\begin{figure*}[t]
    \centering
    \includegraphics[height=0.25\columnwidth,keepaspectratio]{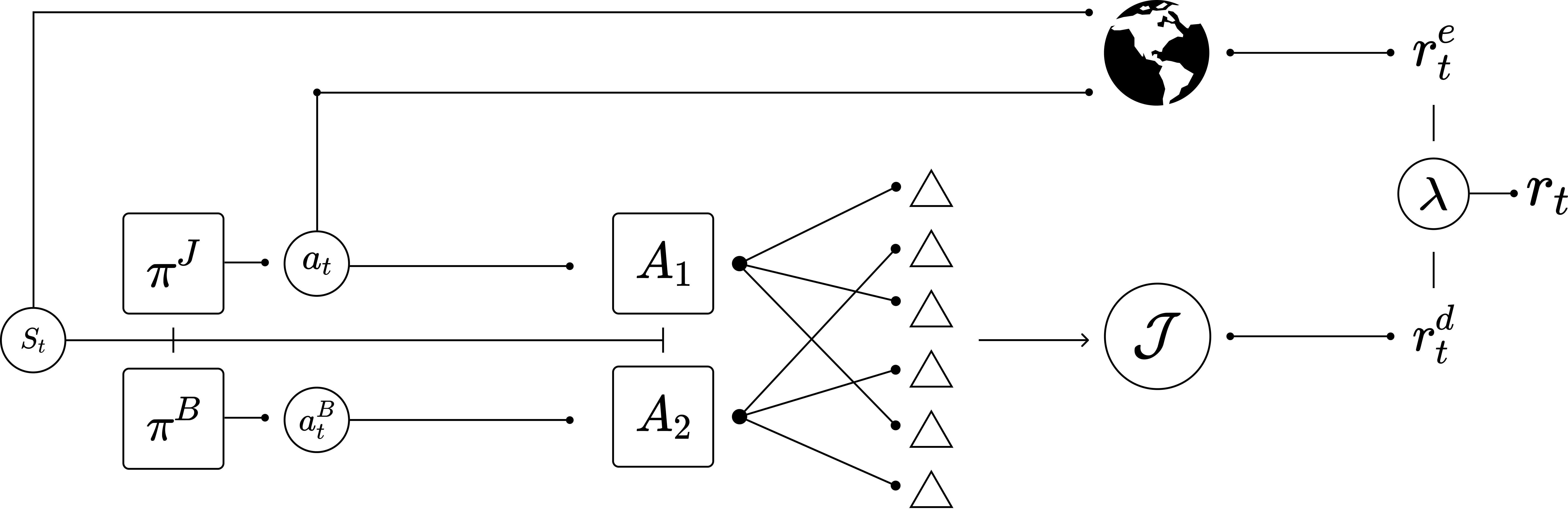} 
    \caption{To obtain a debate reward $r_t^d$ in the state $s_t$, two argumentative agents $A_1$ and $A_2$ take turns proposing supporting evidence (depicted as triangles) for two decisions, up to a predefined limit (here, $3$ evidence each). Then, a positive debate reward is issued whenever a proxy of a judge $\mathcal{J}$ considers action $a_t$, taken by the justifiable policy $\pi^J$, better justified than action $a_t^B$ taken by the baseline policy $\pi^B$. This reward is then mixed with the environment reward $r_t^e$ via debate coefficient $\lambda$, yielding the final reward $r_t$ used to train the justifiable agent.}
    \label{fig:justifiability_framework}
\end{figure*}

To tackle this challenge, we consider a framework which modifies the environment rewards by mixing them with rewards coming from a debate-based reward model (see Figure \ref{fig:justifiability_framework}). Each debate-based reward is defined through the outcome of a two-player zero-sum debate game. More specifically, we let two \textit{argumentative} agents debate by taking turns providing supporting evidence contingent on the current state, each corroborating a decision made by one of two competing policies. Based on the proposed set of evidence, the human then states their preference over these two decisions, and this preference defines the \textit{debate reward}. In this setup, one decision comes from a \textit{baseline} policy, while the other comes from a \textit{justifiable} policy that we optimize. Our approach builds upon the work of \citet{irving2018ai}, but we consider sequential decision-making problems and utilize debate to quantify a decision's justifiability. To this end, we recognize two main technical challenges. First, learning a proxy of a human judge that evaluates decisions solely based on the proposed evidence, with comparable performance to methods requiring full state exposure. Second, learning a representation of argumentative strategies that are solutions to different instances of the debate game. These two components are needed to enable efficient learning of the justifiable policy. 

\partitle{Contributions.} Our contributions are as follows. (i) We formalize the problem of justifiable decision-making, modeling debate as an extensive-form game (Sec. \ref{sec:formal}). (ii) We provide a method for learning a proxy of a human judge that evaluates a decision's justifiability using proposed evidence (Sec. \ref{sec:framework:judge}). (iii) We propose an approach to learning contextualized argumentative strategies that constitute approximate solutions of the debate games (Sec. \ref{sec:framework:agents}). (iv) We conduct extensive empirical evaluation of our approach on a real-world problem of treating sepsis, testing the performance and justifiability of policies trained through our framework (Sec. \ref{sec:exp:trade_off}), as well as the effectiveness and robustness of argumentative agents (Sec. \ref{sec:exp:partial_context}, Sec. \ref{sec:exp:argumentation}, and Sec. \ref{sec:exp:xai}) \footnote{Our code is publicly available at \href{https://github.com/aleksa-sukovic/iclr2024-reward-design-for-justifiable-rl}{github.com/aleksa-sukovic/iclr2024-reward-design-for-justifiable-rl}.}.

\section{Related Work}
\partitle{Debate.} \citet{irving2018ai} first outlined theoretical and practical implications of debate as a method for training aligned agents. Debate has also been used to improve factuality of large language models \citep{du2023improving}, reason in knowledge graphs \citep{hildebrandt2020reasoning}, and aid in Q\&A tasks \citep{perez2019finding}. We build on this line of work by bringing the utility of debate in sequential decision-making problems, where it is used as an approach of quantifying justifiability of made decisions. In addition, by leveraging learning from pairwise preferences as in \citet{christiano2017deep}, our framework enables encoding of human judgments in a form of the preferred evidence that renders a decision justified.

\partitle{Reward Design.} Previous work proposes several approaches that address difficulties of reward design, based on natural language \citep{kwon2023reward, bahdanau2018learning}, rule-based methods \citep{jothimurugan2019composable} and preferences \citep{biyik2018batch,christiano2017deep}. Furthermore, there is a line of work considering interpretable reward design, including reward sparsification \citep{devidze2021explicable} and reward decomposition \citep{juozapaitis2019explainable,bica2020learning}. Differently, we define a reward model as an outcome of a zero-sum debate game, which in itself is interpretable, and learn contextualized policies that solve it. Most similar to our methodology is learning from pairwise preferences as in \citet{christiano2017deep}, only we assume comparisons are made over justifying evidence, without exposure to the underlying state or trajectory. 

\partitle{Explainable AI and Other Related Work.} 
Our approach is most similar to the attribution-based techniques for explaining the inner workings of models. In such approaches, contributions of input features are quantified numerically \citep{lundberg2017unified,ribeiro2016should} or visually represented as heatmaps \citep{selvaraju2017grad,mundhenk2019efficient}. This line of research has received a significant attention, also in the context of explaining decisions of RL agents \citep{ragodos2022protox,bastani2018verifiable}. However, one limitation of these explanations is the inability to further align them with human preferences \citep{hadfield2021explanation, brundage2020toward}. In contrast, our approach additionally enables human-specified justifications, facilitating the incorporation of preferences into their generation. There is also a line of work that examines approaches for adaptation of agent's recommendations (actions) to a baseline human policy in an expert-in-loop setup \citep{grand2022best,faros2023q}. Differently, we consider a problem of learning to take actions that can be justified, where the agent itself acts as a primary decision-maker in the environment.

\section{Formal Setup}\label{sec:formal}
We consider a sequential decision-making problem, where an agent interacts with an environment over a series of time-steps, modeled by a discrete-time Markov Decision Processes (MDP). The episode begins by sampling a starting state from the initial state distribution $\rho$. At each time-step $t$, an agent observes the current state $s_t \in \mathcal{S}$, takes an action $a_t \in \mathcal{A}$ and receives an environment reward $r^e(s_t, a_t)$. The environment then transitions to a successor state $s_{t+1}$ with a probability specified by the transition function $T(s_{t+1} | s_t, a_t)$.

\subsection{Agents}\label{sec:formal:agents}
We consider two kinds of agents that operate in the defined MDP: \textit{baseline} and \textit{justifiable} agent.

\partitle{Baseline Agent.}\label{sec:formal:agents:baseline}
The \textit{baseline agent} aims to maximize the expected discounted value of the environment's return given as $\mathcal{R} = \sum_{t=0}^{T-1} \gamma^t r^e(s_t, a_t)$, where $\gamma \in [0, 1]$ is a user-specified discount factor. Its optimal action-value function, defined as $Q^{\ast}(s, a) = \max_\pi \mathbb{E}\left[ \mathcal{R} | s_t = s, a_t = a, \pi \right]$, is the maximum expected discounted return obtained after taking action $a$ in state $s$. We will refer to a deterministic policy that maximizes the expected value of $\mathcal{R}$ as the \textit{baseline policy} $\pi^B$, which satisfies $\pi^B(s) \in \text{argmax}_{a} \; Q^\ast (s, a)$. \footnote{In practice, we require the baseline policy to be well-performing, but not necessarily optimal.}

\partitle{Justifiable Agent.}\label{sec:formal:agents:justifiable}
While the baseline agent learns to solve the environment, its decisions may not be easy to justify when evaluated by a human. To design an agent that accounts for the justifiability of its decisions, we consider a reward defined as a weighted combination of the environment reward $r^e$ and the \textit{debate reward} $r^d$, which encapsulates human judgment of justifiability and is specified in the next subsection. The expected return is then defined as:
\begin{equation*}
\mathcal{R}_J = \sum\nolimits_{t=0}^{T-1} \gamma^t \left[(1 - \lambda) \cdot r^e(s_t, a_t) + \lambda \cdot r^d(s_t, a_t, a^B_t)\right]
\end{equation*}
where $\lambda \in [0, 1]$ is a \textit{debate coefficient}, balancing between environment and debate rewards, and $a_t^B$ is the action of the baseline agent in state $s_t$. The value of $\lambda=0.0$ corresponds to the return of the baseline policy, whereas for the value $\lambda=1.0$ we obtain a setup similar to \citet{christiano2017deep}, where the agent relies only on the debate reward model. We refer to an agent that maximizes the expected value of $\mathcal{R}_J$ as the \textit{justifiable agent}, denote its optimal action-value function as $Q^{\ast}_{J}(s, a) = \max_\pi \mathbb{E}\left[ \mathcal{R}_{J} | s_t = s, a_t = a, \pi \right]$ and its deterministic policy $\pi^J(s) \in \text{argmax}_{a} Q^{\ast}_J(s, a)$ as the \textit{justifiable policy}.

\subsection{Reward Modeling via Debate}\label{sec:formal:debate}
Our objective is to learn a reward model, denoted as $r^d(s_t, a_t, a_t^B)$, which quantifies the justifiability of a decision. To this end, we introduce a reward model based on a \textit{debate game}. In this model, $r^d(s_t, a_t, a_t^B)$ represents the value of a debate game induced by a tuple $(s_t, a_t, a_t^B)$. In more details, for a given tuple $(s_t, a_t, a_t^B)$, the induced debate game is formulated as a two-player zero-sum extensive-form game \citep{shoham2008multiagent}, in which the first player argues for taking the decision $a_t$ in state $s_t$, while the second player argues for taking the baseline decision $a_t^B$ in $s_t$. 

\partitle{Debate Game.} With $\mathcal{N}$ we denote a set of nodes in a finite, rooted game tree. The action space is represented by a finite set of evidence $\mathcal{E}$, and each node $n \in \mathcal{N}$ consists of evidence proposed thus far in the game $n = \{ e \} \subseteq \mathcal{E}$. Additionally, the edges to successors of each node define actions (evidence) $\{e: e \in \mathcal{E} \setminus n\}$ available to the acting player, where we disallow evidence repetition. The debate game is a perfect-information game: at all times, the players have no ambiguity about the evidence proposed up until the current point and have a complete knowledge about the state of the game. 
The game proceeds as players take turns: in turn $l$, player $i = l \mod 2 + 1$ proposes  evidence $e_{l/2}^{i}$.
\footnote{In our implementation of the debate game, we randomly chose which player has the first turn, i.e., $i = (l + \tau) \mod 2 + 1$, where $\tau \sim \mathcal U(\{0, 1\})$. This only affects the order of the evidence in $n_L$.} 
The total number of turns $L$ is assumed to be even and significantly smaller than the evidence set, i.e., $L \ll |\mathcal{E}|$. After the last turn, a terminal node $n_L = (e_1^1, e_1^2,..., e_{L/2}^1, e_{L/2}^2) = \{ e_{n_L} \}$ is evaluated. The players' utilities are $u_1(n_L) = -u_2(n_L) = \mathbb{U}(a_t, a_t^B, \{ e_{n_L} \})$, with $\mathbb{U}$ defined as:
\begin{equation*}
\mathbb{U}(a_t, a_t^B, \{ e \}) =
    \left
        \{\begin{array}{lr}
            +1, &\;\; \hspace{0.5em} \mathcal{J}(a_t, \{ e \}) > \mathcal{J} (a_t^B, \{ e \}) \\
            0, &\;\; \mathcal{J}(a_t, \{ e \}) = \mathcal{J}(a^B_t, \{ e \})\\
            -1, & \text{otherwise}
        \end{array}
    \right.
\end{equation*}
Here, $\mathcal{J}$ is a model of a human judge that, for a given decision $a$ and evidence $\{ e \}$, outputs a numerical value $\mathcal{J}(a, \{ e \}) \in \mathbb{R}$ quantifying how justifiable $a$ is under evidence $\{ e \}$.

\partitle{Strategies and Solution Concept.}
A player's strategy $\sigma^i: \mathcal{N} \to \mathcal{E}$ outputs available evidence $\sigma^i(n) \in \mathcal{E} \setminus n$ in a given node $n$ and $\Sigma^i$ is the set of all strategies of the player $i$. Based on the utility function, we additionally define $\mathcal{G}(\{\sigma^1, \sigma^2\}, s_t, a_t, a_t^B)$ as the \textit{payoff} (utility) of the first player, conditioned on both players following the strategy profile $\{\sigma^1, \sigma^2\}$. Then, a set of the best responses of the first (resp. second) player to its opponent strategy $\sigma^{2}$ (resp. $\sigma^{1}$) is defined as $b^1(\sigma^{2}) = \arg \max_{\sigma^1 \in \Sigma^1} \mathcal{G}(\{\sigma^1, \sigma^{2}\}, s_t, a_t, a_t^B)$ (resp. $b^2(\sigma^{1}) = \arg \min_{\sigma^2 \in \Sigma^2} \mathcal{G}(\{\sigma^1, \sigma^{2}\}, s_t, a_t, a_t^B)$). A strategy profile $\bar \sigma = \{\bar \sigma^1, \bar \sigma^2\}$ is a pure-strategy Nash equilibrium if $\bar\sigma^i \in b^i(\bar \sigma^{-i})$. Because the debate game is a perfect-information extensive-form game, a pure-strategy Nash equilibrium exists \citep{shoham2008multiagent}, and due to its zero-sum structure, it can be obtained by solving the following max-min optimization problem:    
$
    \max_{\sigma^1 \in \Sigma^1} \min_{\sigma^2 \in \Sigma^2} \mathcal{G}(\{\sigma^1, \sigma^2\}, s_t, a_t, a_t^B)
$. We refer to $\mathcal{G}(\{\bar \sigma^1, \bar \sigma^2\}, s_t, a_t, a_t^B)$ as the value of the game \footnote{For a two-player zero-sum game, the value of the game (or payoff) is unique \citep{von1947theory}.} and define $r^d(s_t, a_t, a_t^B)$ to be equal to it, i.e., $r^d(s_t, a_t, a_t^B) = \alpha \cdot \mathcal{G}(\{\bar \sigma^1, \bar \sigma^2\}, s_t, a_t, a_t^B)$, where $\alpha > 0$ is a scaling coefficient.

\section{Learning Framework}\label{sec:framework}
To effectively use a debate game outcome during training of justifiable policies, it is necessary to devise a model of a human judge $\mathcal{J}$ (Sec. \ref{sec:framework:judge}) that encapsulates justifiability judgment and additionally learn argumentative policies that approximate a Nash equilibrium and are able to generalize across different instances of the debate game. With $\hat{r}^d(s_t, a_t, a_t^B)$ we denote an approximation of $r^d(s_t, a_t, a_t^B)$, obtained by running the argumentative policies from Sec. $\ref{sec:framework:agents}$ in the debate game induced by $(s_t, a_t, a_t^B)$. In all our experiments, we set $\alpha=5$.

\subsection{Preference Dataset}\label{sec:framework:dataset}
We assume the human judgments are collected in a preference dataset $\mathcal{D}$ of tuples $(s_t, a_0, a_1, p)$, where $s_t$ is a state, $a_0 \ne a_1$ are two decisions and $p \in \{0, 1\}$ indicates which of the two decision is more justified in a particular state. The value of $p=0$ (resp. $p=1$) indicates that $a_0$ (resp. $a_1$) is more preferred.

\subsection{Judge Model}\label{sec:framework:judge}
Because asking for human feedback is expensive, we aim to learn a judge model from the dataset $\mathcal{D}$ of preferences that can be used to evaluate the outcome of debate games. The judge, parametrized with learnable parameter $\theta \in \mathbb{R}^{d_1}$, is defined as a scalar reward function $\mathcal{J}_{\theta}(a, \{ e \}) \in \mathbb{R}$ quantifying how justifiable a decision $a$ is, given evidence $\{ e \}$. Similar to \citet{christiano2017deep}, we additionally assume that justifiability preference for decision $a_0$ over decision $a_1$ follows the Bradley-Terry 
model \citep{bradley1952rank}: 
\begin{equation*}
\mathcal{P}(a_0 \succ a_1, \{ e \}) = \frac{\exp \mathcal{J_\theta}(a_0, \{ e \})}{\exp \mathcal{J_\theta}(a_0, \{ e \}) + \exp \mathcal{J_\theta}(a_1, \{ e \})}.
\end{equation*}
Here, we require the judge to quantify the level of justifiability given all evidence at once. Note the lack of dependency on the state $s_t$: the judge evaluates only proposed evidence, whereas the argumentative agents are in charge of providing those evidence, contingent on the state. We optimize the parameters by minimizing the cross-entropy loss between preference-predictions and labels from the dataset. See App. \ref{appendix:models:judge} for more details.

\subsection{Argumentative Agent}\label{sec:framework:agents}
Our overarching goal is to learn a generalizable argumentative policy that is able to solve any debate game, conditioned on its defining tuple $(s_t, a_t, a_t^B)$. This is a difficult feat, as the evidence set available to the agent is contingent on the state $s_t$. To this end, we can treat the debate game as an instance of a contextualized extensive form game \citep{sessa2020contextual}, where we consider the tuple $(s_t, a_t, a_t^B)$ as a context $z \in \mathcal{Z}$. In a general case, the justifiable agent $\pi^J$ observes the state $s_t$ and takes action $a_t$ which, paired with an action $a_t^B$ the baseline agent $\pi^B$ would have taken, sets the debate game context $z = \{ s_t, a_t, a_t^B \}$. In the specific case of offline reinforcement learning we consider here, the contexts are sampled i.i.d. from the static dataset throughout training of both, argumentative and justifiable agents. Therefore, given a sample from the preference dataset $(s_t, a_0, a_1, p) \sim \mathcal{D}$, we set the context to $z=(s_t, a_p, a_{1-p})$. 
Player $i$'s contextual strategy, parametrized with learnable parameter $\phi_i \in \mathbb{R}^{d_2}$, is defined as $\sigma^c_{\phi_i}: \mathcal{Z} \to \Sigma^i$, 
mapping a context to the strategy of the player for the induced debate game \footnote{Equivalently, we will also denote the contextual strategy of player $i$ as $\sigma^c_{\phi_i}(\cdot | z): \mathcal{N} \to \mathcal{E}$ for $z \in \mathcal{Z}$.}. 
To learn parameters $\phi_i$, we solve:
$
    \max_{\phi_1 \in \mathbb{R}^{d_2}} \min_{\phi_2 \in \mathbb{R}^{d_2}} \mathbb{E}_{z \sim \mathcal{D}} [\mathcal{G}(\{ \sigma^c_{\phi_1}(z), \sigma^c_{\phi_2}(z) \}, z)]
$.

\subsection{Method}\label{section:method}

\partitle{Judge.} The judge $\mathcal{J}_\theta$ is parametrized by weights $\theta \in \mathbb{R}^{d_1}$ of a neural network with two fully-connected layers of size $256$, using parametric relu \citep{he2015delving} activation and batch normalization \citep{ioffe2015batch}. The network receives a vector $x$ in $\mathbb{R}^{|\mathcal{E}|}$, where only the values of evidence $\{e\}$ are shown, while the rest are set to zero. In addition, a binary mask of the same dimension, wherein all elements corresponding to the evidence $\{e\}$ are assigned a value of one, while the remaining elements are set to zero, as well as a one-hot encoded action $a$ are passed. The learning is done for a total of $100$ epochs using batches of $64$ comparisons sampled from the preference dataset $\mathcal{D}$, Adam optimizer and a learning rate of 5e\text{-}4. See App. \ref{appendix:models:judge} for more details.

\partitle{Argumentative Agent.}
We represent parameters $\phi \in \mathbb{R}^{d_2}$ of the argumentative agent $\sigma^c_{\phi}(\cdot | z)$, as weights of a neural network composed of $2$ fully-connected layers of size $512$ with leaky-relu activation function \citep{maas2013rectifier} with slope of $1\text{e-}2$. The network takes as input a $44$-dim state, a one-hot encoded decision for which the agent argues, as well as a binary mask of currently proposed evidence. The evidence (action) space of the policy is discrete and has $44$ choices, each corresponding to exactly one state feature. To train the agent, we use PPO \citep{schulman2017proximal} and examine two optimization strategies, namely \textit{self-play} and \textit{maxmin}. We train the \textit{self-play} debate by letting the agent argue with a copy of itself \footnote{In terms of learnable parameters, this implies $\phi_2 \coloneqq \phi_1$.}, updating that copy every $100$k steps, and repeating the procedure for $500$ generations. For \textit{maxmin} debate, we use $2$ different set of weights, $\phi_1$ and $\phi_2$, to represent the agents' policies. The first agent i.e., $\phi_1$ is trained for $4$k steps, followed by training of the second agent i.e., $\phi_2$ for $100$k steps, repeating this procedure for $500$ generations. This approach of overfitting the second agent to the current version of the first ensured learning of defense strategies against a very strong adversary. See App. \ref{appendix:models:argumentation} for more details.

\partitle{Justifiable Agent.}
To learn a justifiable policy $\pi^J(s) \in \argmax_a Q^\ast_J(s, a)$, we consider model-free reinforcement learning methods based on Q-learning. Our approach builds on top of \citet{raghu2017deep} and is based on a variant of Deep-Q networks \citep{mnih2015human}, specifically double-deep Q-network with the dueling architecture \citep{wang2016dueling,van2016deep}. The final network consists of $2$ fully-connected layers of size $128$ using leaky-relu activation function with slope $1\text{e-}2$. The learning is done in batches of $256$ $(s, a, r, s')$ tuples sampled from a Prioritized Experience Replay buffer \citep{schaul2015prioritized} using a learning rate of $1\text{e-4}$, for a total of $25$k iterations, evaluating the policy every $50$ iterations on a held-out test set. When incorporating signal from the debate, i.e., $\hat{r}^d$, we augment the reward from the replay buffer as described in Sec. \ref{sec:formal:agents:justifiable}. Modifying just the observed reward, without changing the state dynamics, has been shown to be sufficient to induce learning of an alternative policy \citep{ma2019policy}. See App. \ref{appendix:models:sepsis} for more details, including additional loss terms and a list of all hyperparameters.

\section{Experiments}\label{sec:exp}
In our experiments, we aim to empirically evaluate the properties of debate as a method for reward specification. To this end, we perform quantitative and qualitative evaluation of justifiable policies (Sec. \ref{sec:exp:trade_off}), analyze the effect of pairwise comparison only over proposed evidence on performance and justifiability (Sec. \ref{sec:exp:partial_context}), examine the necessity of multi-agent debate in learning to provide evidence that is resilient to refutations (Sec. \ref{sec:exp:argumentation}), and compare efficiency in providing supporting evidence for a decision of argumentative policies and SHAP feature-attribution method  (Sec. \ref{sec:exp:xai}).

\begin{figure*}
    \centering
    \begin{subfigure}[c]{0.8\textwidth}
        \centering
        \includegraphics[width=\textwidth,keepaspectratio]{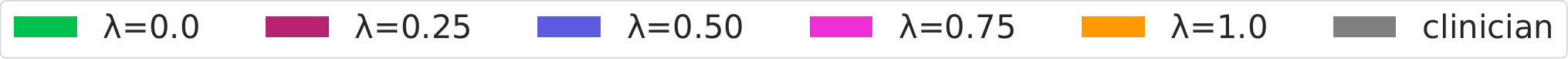}
    \end{subfigure}%
    \hfill%
    \\[1.5ex]
    \begin{subfigure}[b]{0.24\textwidth}
        \centering
        \includegraphics[width=\textwidth]{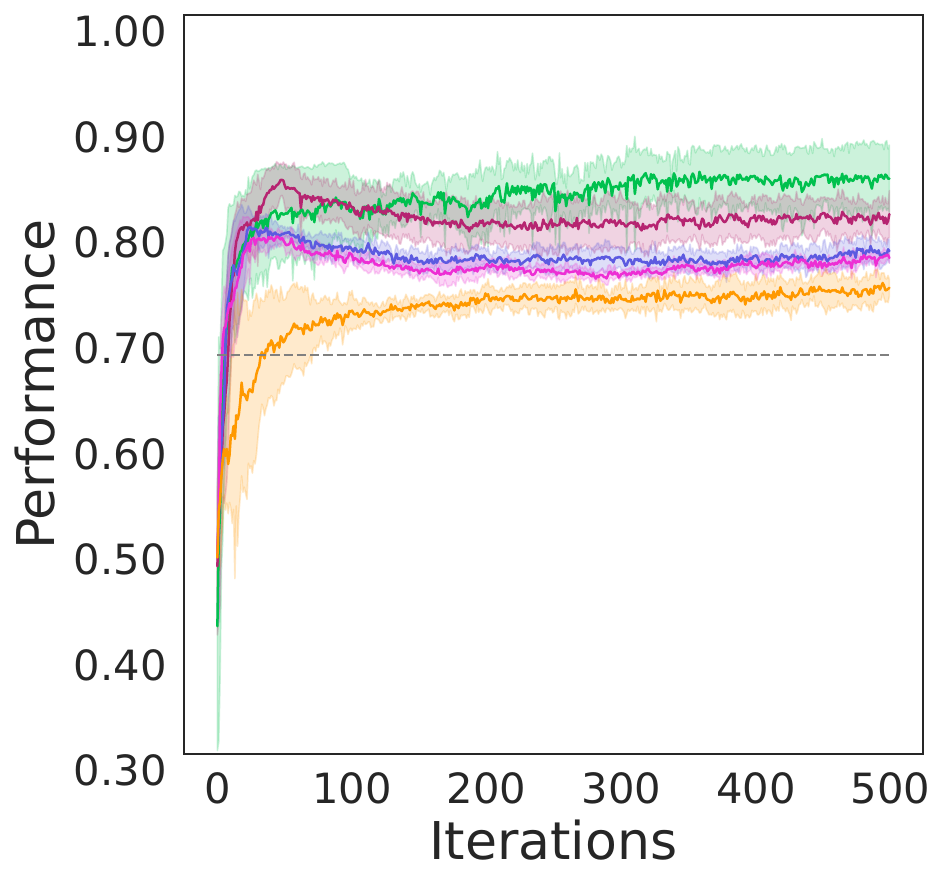}
        \caption{WIS evaluation}
        \label{plt:wis_evaluation}
    \end{subfigure}
    \hfill
    \begin{subfigure}[b]{0.24\textwidth}
        \centering
        \includegraphics[width=\textwidth]{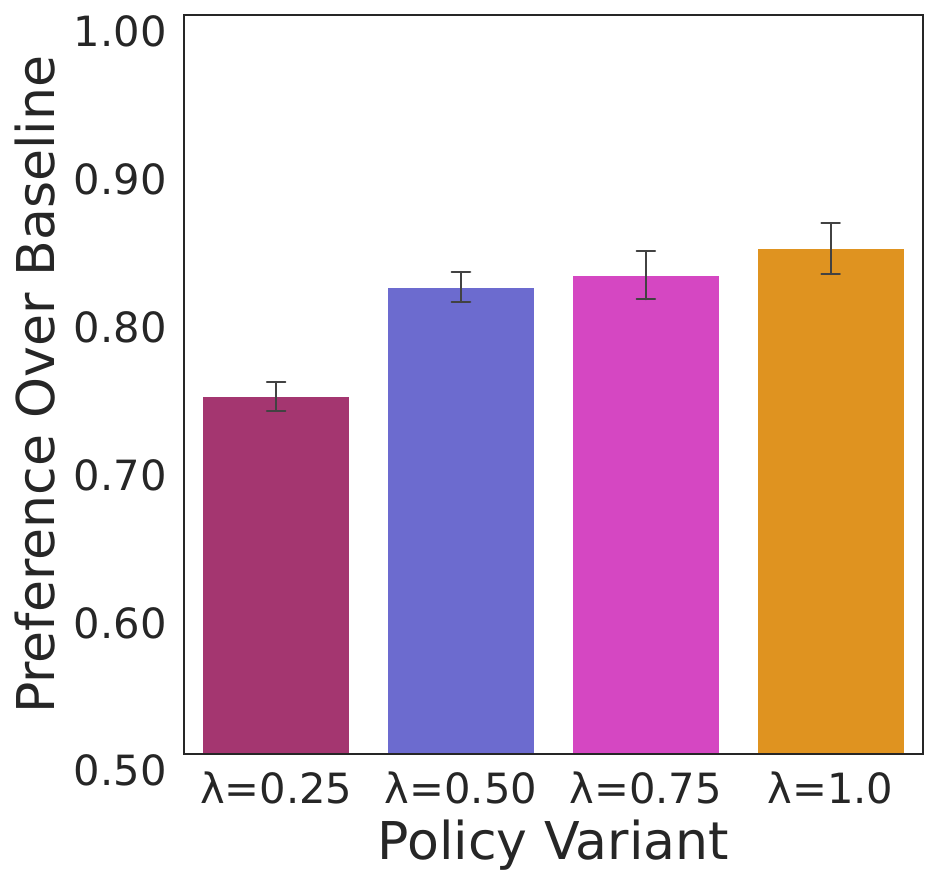}
        \caption{Preferred to baseline}
        \label{plt:jstf_evaluation}
    \end{subfigure}
    \begin{subfigure}[b]{0.24\textwidth}
        \centering
        \includegraphics[width=\textwidth]{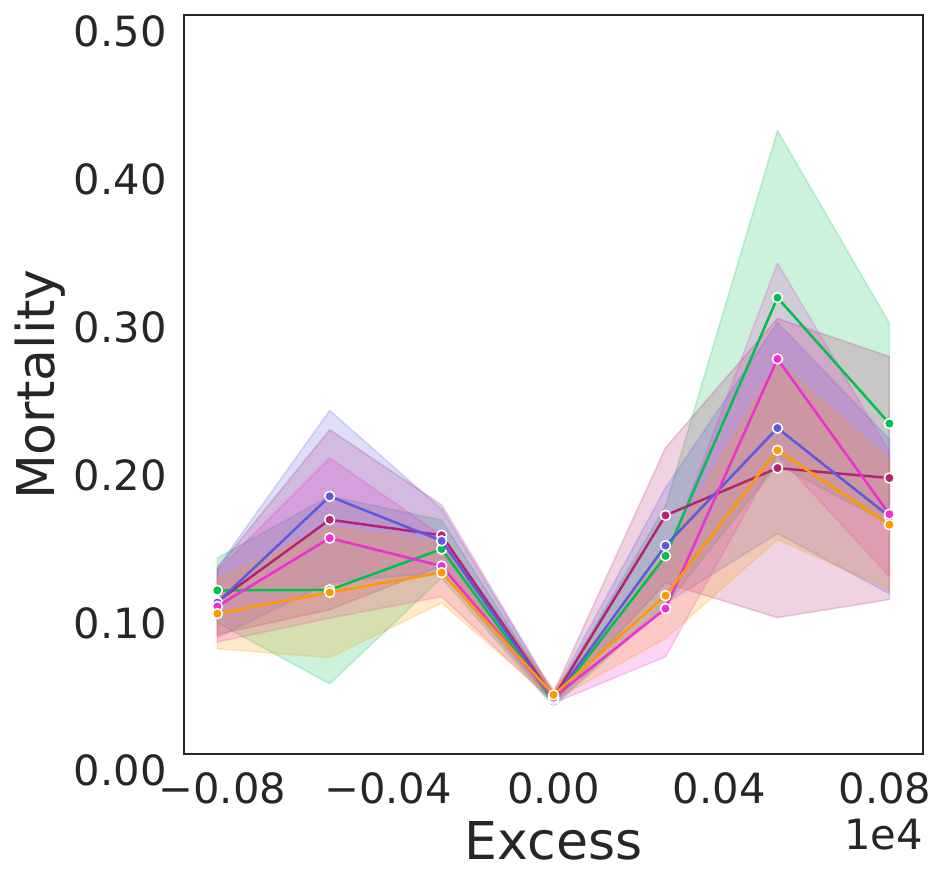}
        \caption{IV dose excess}
        \label{plt:iv_dose_excess}
    \end{subfigure}
    \hfill
    \begin{subfigure}[b]{0.24\textwidth}
        \centering
        \includegraphics[width=\textwidth]{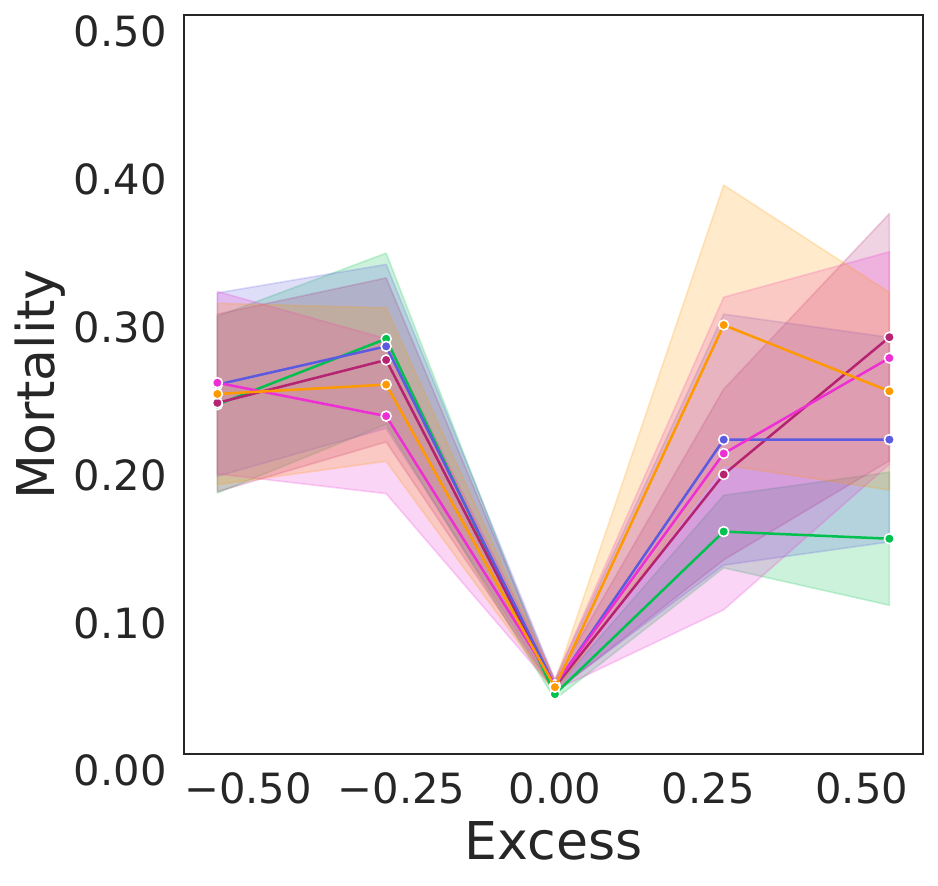}
        \caption{VC dose excess}
        \label{plt:vc_dose_excess}
    \end{subfigure} 
    \caption{Evaluation of justifiable policies. For (b)-(d), the confidence intervals represent $\pm 2$ standard errors of the mean over $5$ random seeds. (a) Policy performance as measured by WIS evaluation on a held-out test set with $\pm 1$ terminal rewards for every patient discharge or death. The mean and standard deviation are reported over $5$ random seeds. (b) Percent of times judge preferred decisions of justifiable policies (i.e., $\lambda > 0.0$) compared to those of the baseline policy (i.e., $\lambda=0.0$). (c) (d) Observed patient mortality (y-axis) against variations in IV/VC treatment doses prescribed by clinicians compared to the recommendations of learned policies (x-axis).}
    \label{fig:protagonist_evaluation}
\end{figure*}

\subsection{Environmental Setup}\label{sec:exp:environment}
\partitle{Sepsis.}\label{sec:exp:environment:sepsis} Data for our cohort were obtained following steps outlined in \citet{komorowski2018artificial}, utilizing MIMIC-III v1.4 database \citep{johnson2016mimic}. We focus our analysis on patients that fulfill Sepsi-3 criteria \citep{singer2016third}, 18,585 in total. The patient vector consists of $40$ continuous and $4$ discrete features, and the action space consists of $5 \times 5$ discrete choices of intravenous (IV) fluids and vasopressors (VC). As in \citet{raghu2017deep}, the environment rewards are clinically guided. We set the environment reward $r^e$ to $\pm 15$ for terminal states resulting in patient survival or death, respectively, and shape the intermediate rewards based on the SOFA score (measure of organ failure). See App. \ref{appendix:dataset} for more details.

\partitle{Preference Dataset.}\label{sec:exp:environment:dataset} To make a comprehensive evaluation possible, we define a synthetic ground-truth preference by making an assumption that a human judge always prefers a treatment prescribed by the clinician. Therefore, clinician's actions are the justified actions in our experiments. More formally, we bootstrap the dataset of preferences $\mathcal{D}$ by matching every pair $(s_t, a_t)$ from the cohort with an alternative action $a_r \sim \mathcal{U}(A), \; a_r \ne a_t$ sampled uniform-random from the action space, initializing the preference variable $p$ to point to the true action $a_t$, as taken by the clinician (see App. \ref{appendix:additional_results:dataset} for alternative dataset bootstrapping methods). The number of evidence is fixed to $6$ ($\sim$ 13.6\% of the full state) (see App. \ref{appendix:additional_results:argumentation} for results with $L=4$ evidence). The dataset is split into chunks of 70\%, 15\%, 15\% used for training, validation, and testing respectively. We report all our results on a held-out test set, not seen by any of the agents nor the judge. When training a judge model, we additionally augment a tuple $(s_t, a_0, a_1, p)$ with an evidence set $\{ e \}$. To generate it, we sample a predetermined number of state features uniform-random i.e., $\{ e \} \sim \mathcal{U}(s_t)$, from the state, which is inspired by \citet{irving2018ai} and ensures evidence is contingent on the state (Sec. \ref{sec:formal:debate}) \footnote{We discuss different approaches for defining a set of evidence in Sec. \ref{sec:discussion:arguments}.}. On the test set, the judge is able to correctly recover the underlying preference from the dataset with a relatively low accuracy of $65\%$.

\partitle{Baselines.}\label{sec:exp:environment:baselines} When comparing effectiveness of treatment policies, we consider two baselines. First, we consider the observed reward of the clinician from the dataset (depicted as a gray horizontal line in plots). Second, the baseline policy (Sec. \ref{sec:formal:agents:baseline}) serves as an indicator of the optimal treatment policy. To demonstrate the robustness of multi-agent debate, we introduce an \textit{isolated} argumentative agent. This agent aims to find an evidence set $\{ e \}$ that maximizes $\mathcal{J}(a_p, \{ e \})$, for a given $a_p$. To achieve this, we solve a search problem akin to the debate game by applying reinforcement learning (see App. \ref{appendix:models} for more details). Lastly, we use SHAP when comparing effectiveness of debate to a feature-importance approach in providing supporting evidence for a decision.

\subsection{Experiment 1: Effectiveness of Tasks Policies}\label{sec:exp:trade_off}
To examine the potential of specifying the reward as an adversarial game and its effect on the quality of learned behaviors, we train several policies by varying the debate coefficient $\lambda$.

\partitle{Quantitative Evaluation.} In Plot \ref{plt:wis_evaluation}, we evaluate the performance of different justifiable policies on a held-out test set during the course of training using weighted importance sampling (WIS) \citep{sutton2018reinforcement} \footnote{The behavior policy used in WIS was derived via behavior cloning, further described in App. \ref{appendix:models:sepsis}.}. Likewise, in Plot \ref{plt:jstf_evaluation}, we show the judge's preference over decisions made by justifiable policies (i.e., $\lambda > 0.0$), compared to the baseline policy (i.e., $\lambda=0.0$), when the two were different \footnote{See Plot \ref{plt:appendix:preference_breakdown} in App. \ref{appendix:additional_results} for a detailed breakdown of actions proposed by justifiable policies.}. The observed inherent trade-off between performance and justifiability suggests that tuning the debate coefficient $\lambda$ is important in practice, and we further elaborate on this in Sec. \ref{sec:discussion}. 

\partitle{Qualitative Evaluation.} In addition to quantitative evaluation, we perform qualitative analysis similar to \citet{raghu2017deep}. Plots \ref{plt:iv_dose_excess} and \ref{plt:vc_dose_excess} showcase correlation between observed mortality and difference between the optimal doses suggested by policies, and the actual doses prescribed by the clinicians. For all trained policies, the lowest mortality is observed when the difference is near zero, thus further showcasing their potential validity. It is also encouraging to see that the policy trained solely with debate rewards (i.e., $\lambda=1.0$) remains quantitatively and qualitatively competitive in addition to being highly favored by the judge, even though it relies only on debate rewards.

\begin{figure*}
    \centering
    \begin{subfigure}[b]{0.24\textwidth}
        \centering
        \includegraphics[width=\textwidth]{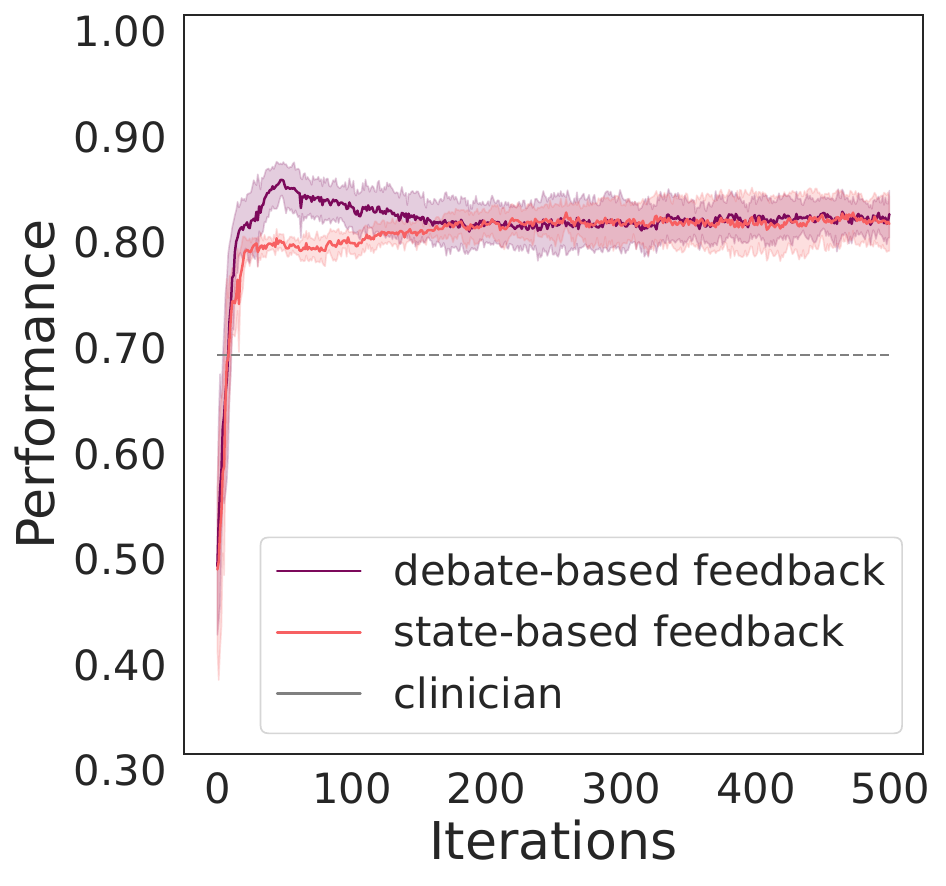}
        \caption{$\lambda=0.25$}
        \label{plt:full_vs_partial_l25}
    \end{subfigure}
    \begin{subfigure}[b]{0.24\textwidth}
        \centering
        \includegraphics[width=\textwidth]{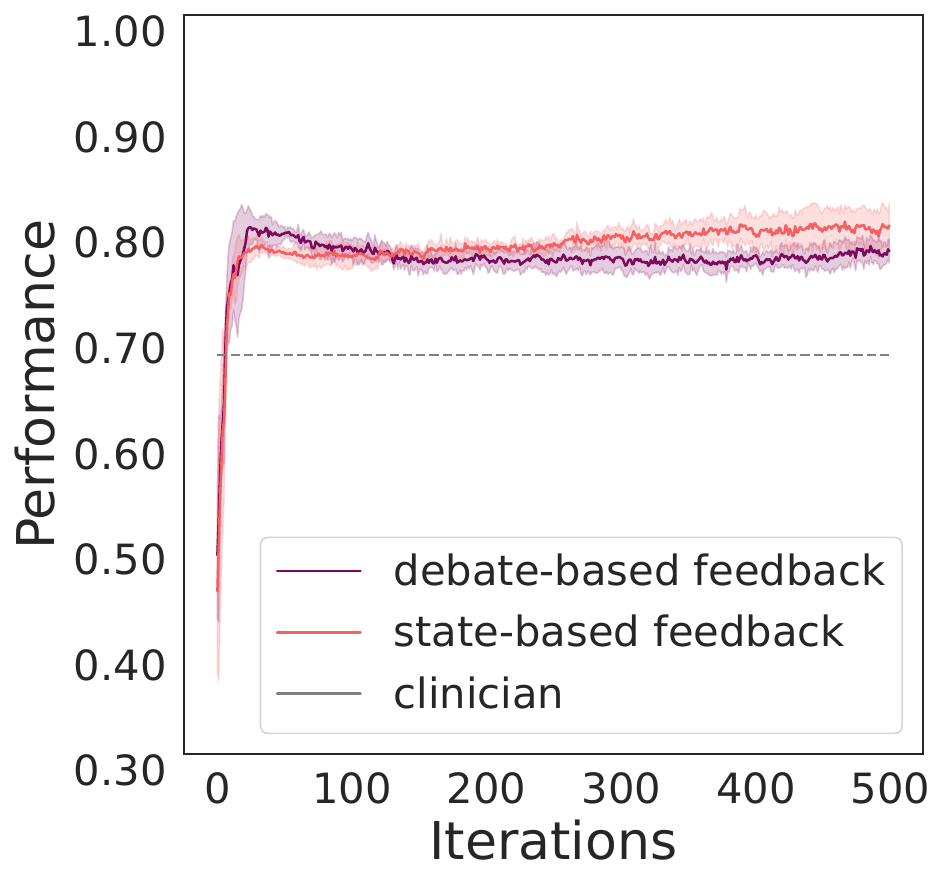}
        \caption{$\lambda=0.50$}
        \label{plt:full_vs_partial_l50}
    \end{subfigure}
    \begin{subfigure}[b]{0.24\textwidth}
        \centering
        \includegraphics[width=\textwidth]{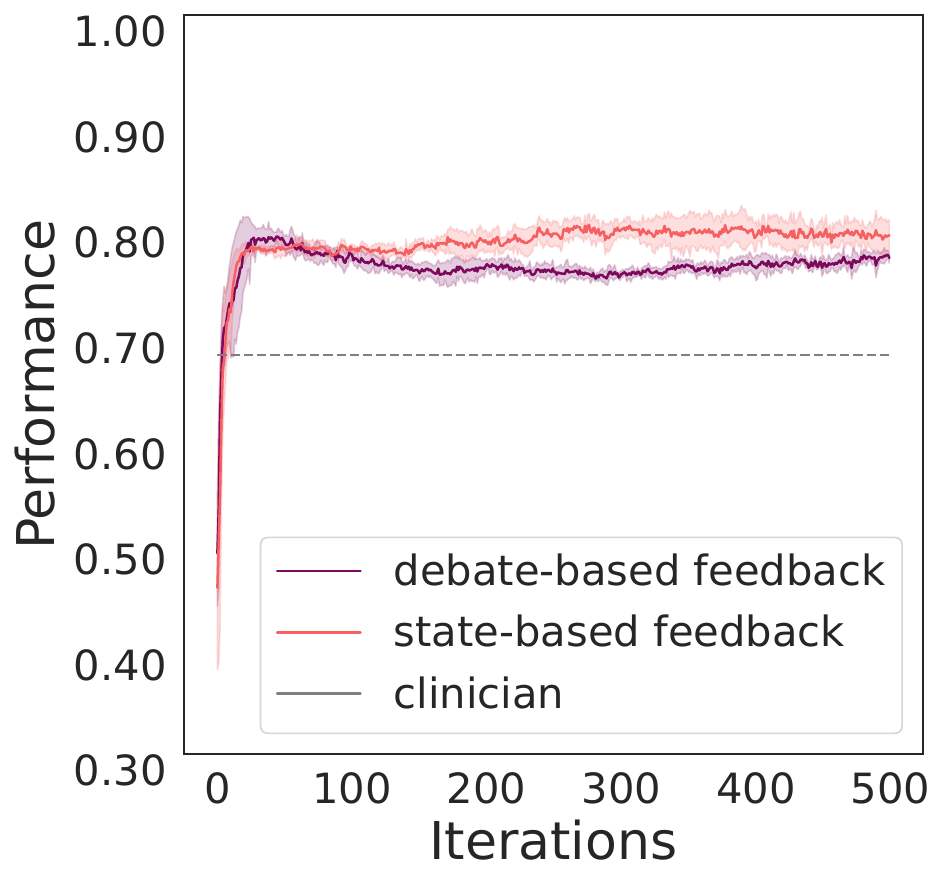}
        \caption{$\lambda=0.75$}
        \label{plt:full_vs_partial_l75}
    \end{subfigure}
    \begin{subfigure}[b]{0.24\textwidth}
        \centering
        \includegraphics[width=\textwidth]{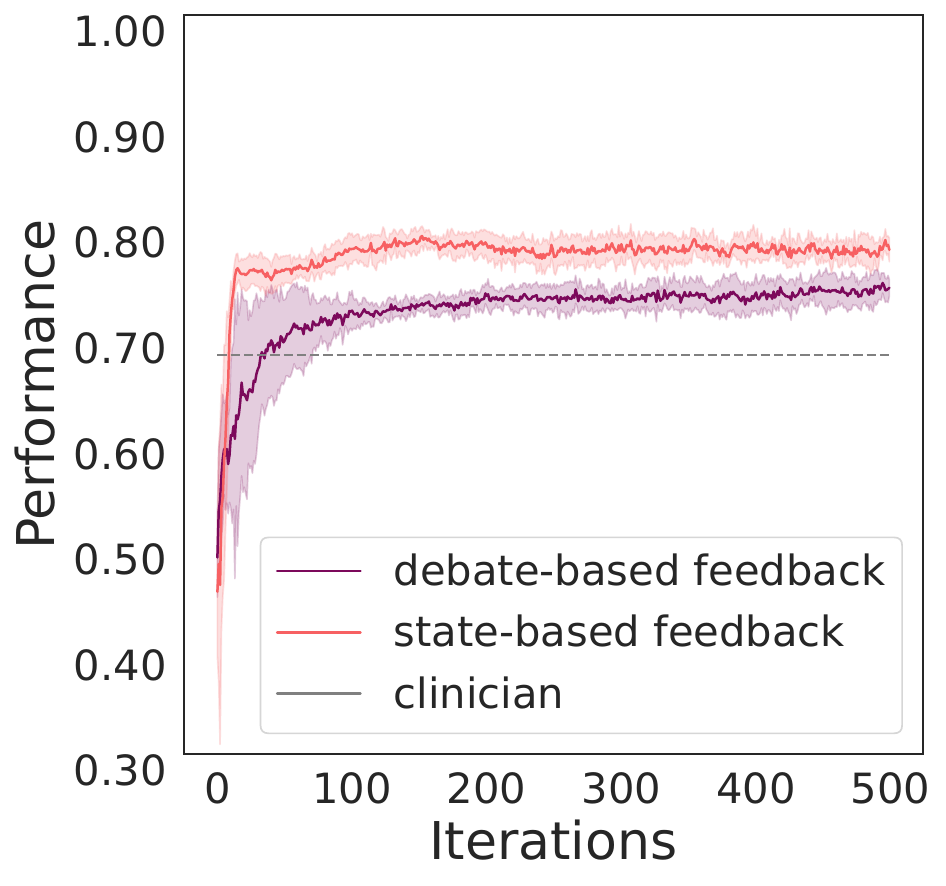}
        \caption{$\lambda=1.0$}
        \label{plt:full_vs_partial_l100}
    \end{subfigure}
    \caption{Performance of policies trained with state-based feedback compared to debate-based feedback, as measured by the weighted importance sampling evaluation on a held-out test set with $\pm 1$ terminal rewards for every patient discharge or death. The mean and standard deviation are reported over $5$ random seeds.}
    \label{fig:full_vs_partial_context}
\end{figure*}

\subsection{Experiment 2: Debate-Based Feedback vs. State-Based Feedback}\label{sec:exp:partial_context}
Because the judge is evaluating justifiability using only proposed evidence, a natural question to ask is how much does this affect the performance and alignment of trained policies. To provide an answer, we train a new judge that evaluates decisions based on the full state, namely $\mathcal{J}^{\prime}(a_t, s_t)$. We then use this judge instead of $\hat{r}^d$ to train a new justifiable policy using \textit{state-based} feedback, one for each $\lambda$. Given a sample $(s_t, a_0, a_1, p) \sim \mathcal{D}$, we consider a policy $\pi$ aligned to the ground-truth preference if $Q^\ast_{\pi}(s, a_p) \ge Q^{\ast}_{\pi}(s, a_{1 - p})$.

The results are shown in Plot \ref{plt:full_vs_partial_context} for various policies trained with state- and debate-based feedback. We see that, even though the debate-based approach uses only $\sim 13.6\%$ of the full state, the achieved level of alignment remains similar. Likewise, in Plots \ref{plt:full_vs_partial_l25}-\ref{plt:full_vs_partial_l100} we observe a matching trend when it comes to performance of the policies, although here we additionally note that the difference seems to increase in favor of policies trained with the state-based feedback as we increase the debate coefficient $\lambda$. These results seem promising, as they indicate one can expect to obtain a competitive level of alignment and performance, while requiring the judge to elicit preference over relatively small number of evidence.

\subsection{Experiment 3: Effectiveness of argumentative policies}\label{sec:exp:argumentation}
\begin{figure*}
    \centering
    \begin{subfigure}[b]{0.24\textwidth}
        \includegraphics[width=\textwidth]{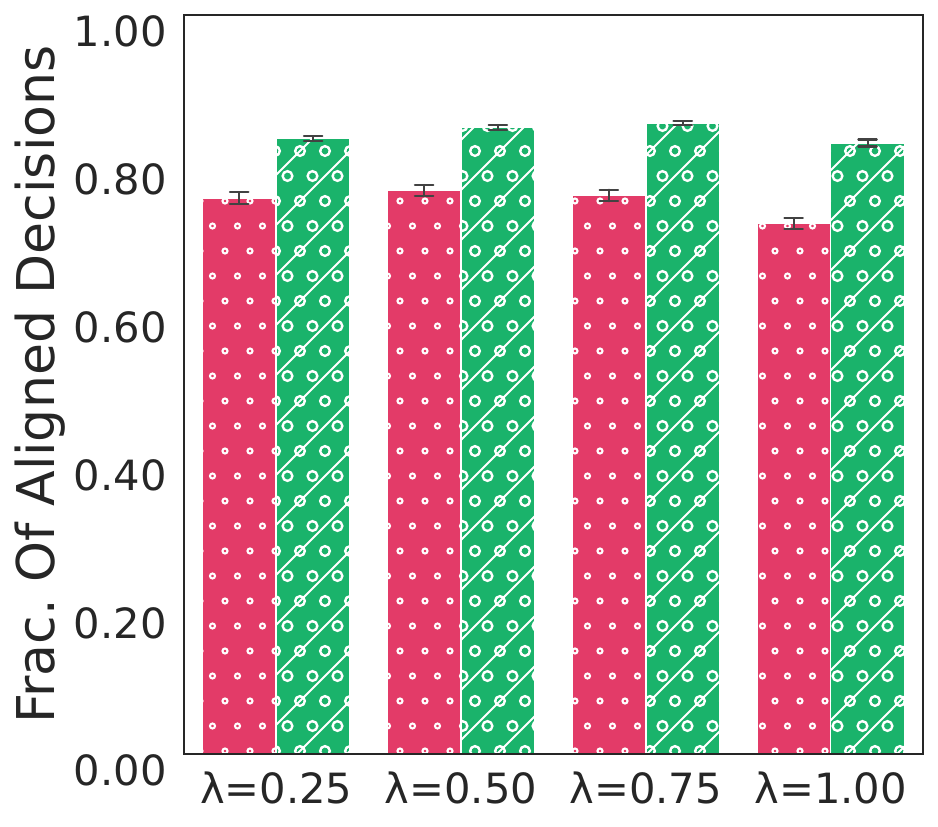}
        \caption{State/debate feedback}
        \label{plt:full_vs_partial_context}
    \end{subfigure}
    \hfill
    \begin{subfigure}[b]{0.24\textwidth}
        \includegraphics[width=\textwidth]{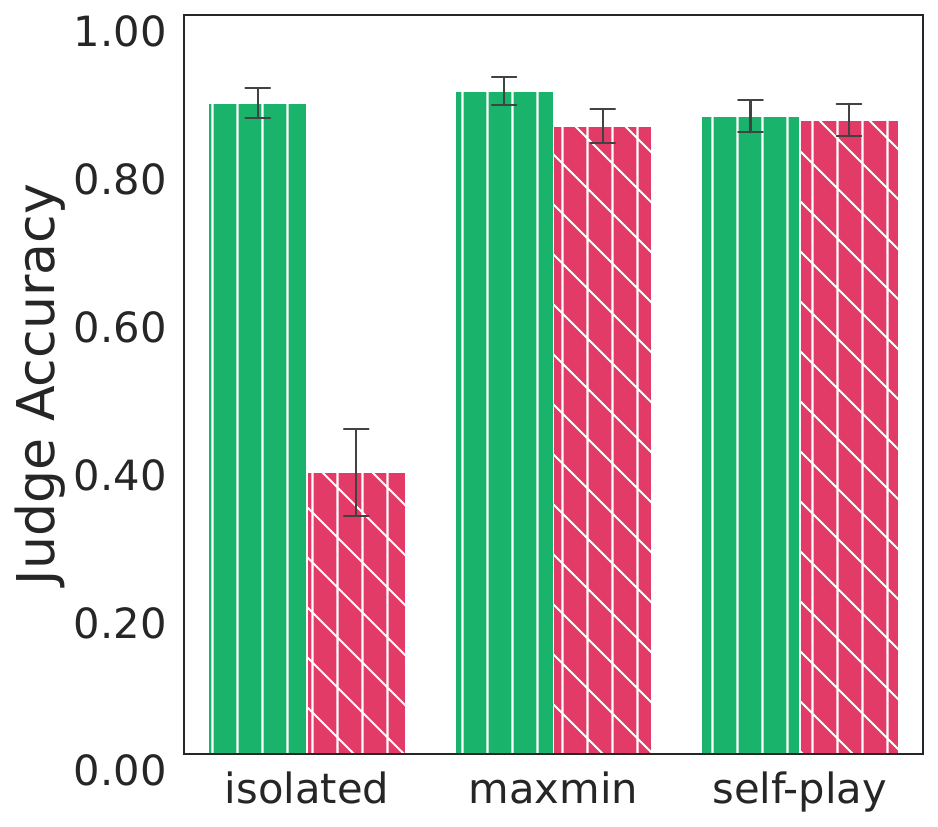}
        \caption{Preference recovery}
        \label{plt:argumentation_confuser}
    \end{subfigure}
    \hfill
    \begin{subfigure}[b]{0.24\textwidth}
        \includegraphics[width=\textwidth]{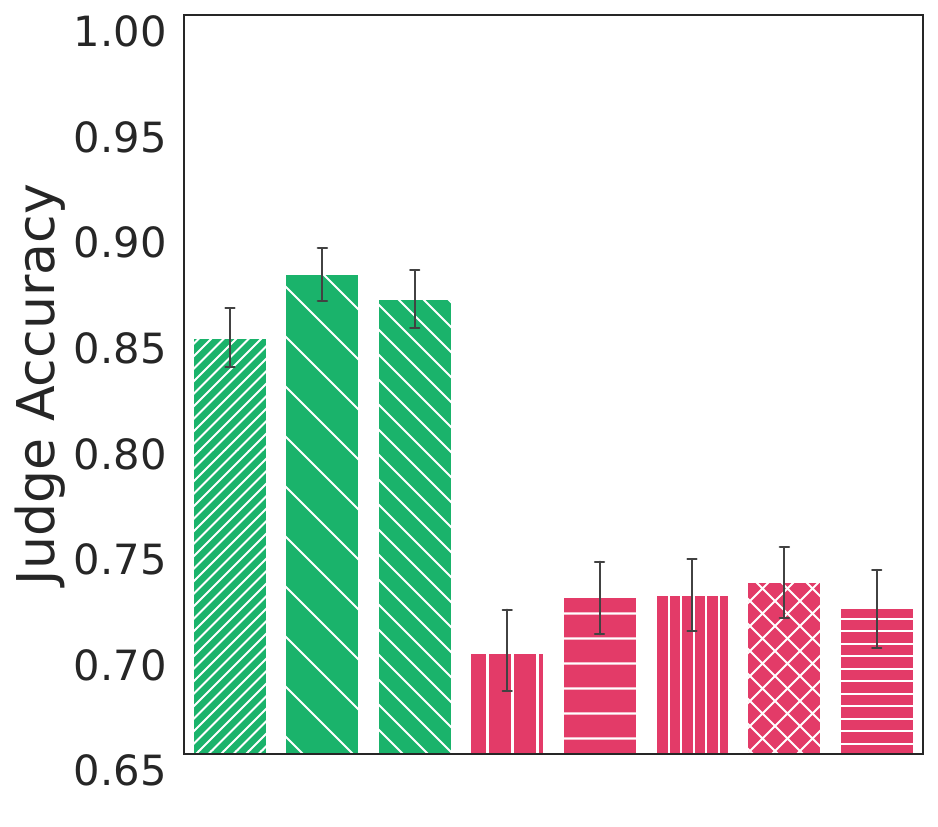}
        \caption{Comparison to SHAP}
        \label{plt:shap}
    \end{subfigure}
    \hfill
    \begin{subfigure}[b]{0.24\textwidth}
        \centering
        \includegraphics[width=\textwidth]{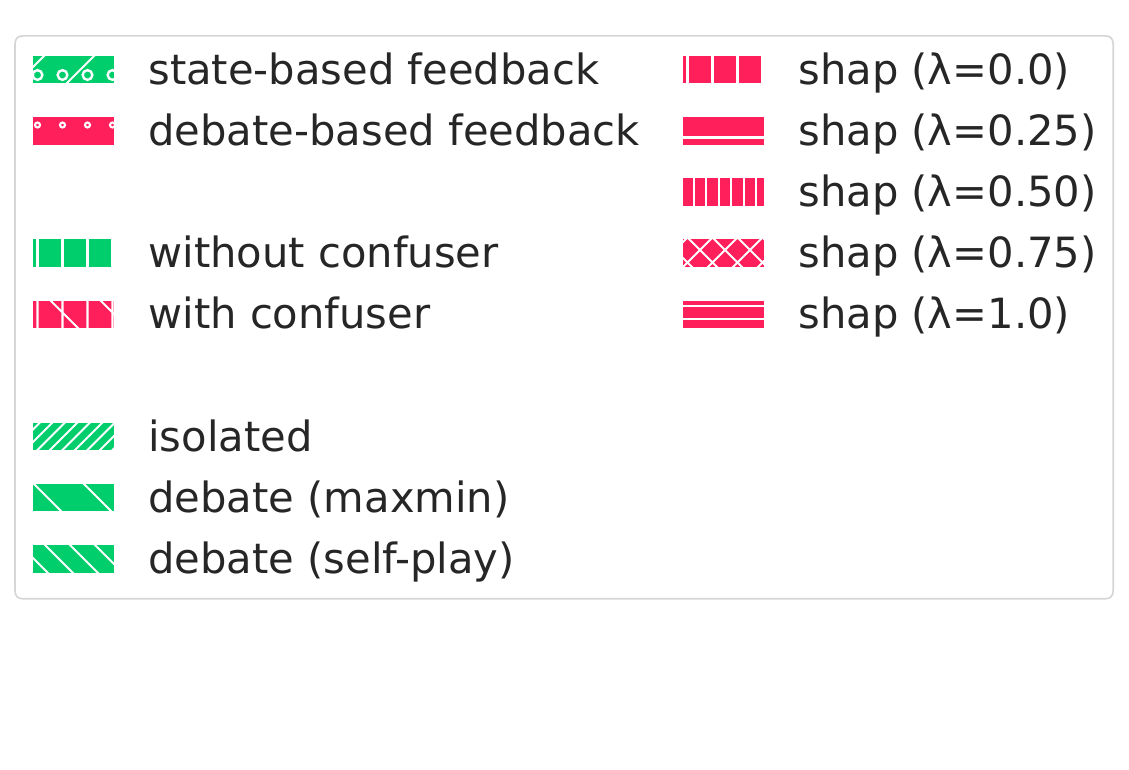}
        \vspace{9.0mm}
    \end{subfigure} 
    \caption{(a) Fraction of aligned decisions of policies trained with debate- and state-based feedback. Confidence intervals (CI) represent $\pm 2$ standard errors of the mean over $5$ random seeds. (b) Accuracy of the judge in predicting the preferred action, with and without the confuser agent, with CI representing $\pm 2$ standard errors of the mean estimate. (c) Effectiveness of SHAP-based explanations when used to justify a decision, as measured by the judge's accuracy, with CI representing $\pm 2$ standard errors of the mean estimate.}
\end{figure*}

\partitle{Preference Recovery Rate.}\label{sec:exp:argumentation:prr} We recall the judge's accuracy in correctly predicting the preferred action $a_p$ from the preference dataset was $65\%$. To evaluate the effectiveness of argumentative policies, for each sample $(s_t, a_0, a_1, p) \sim \mathcal{D}$ we measure the judge's accuracy in predicting the more justified action $a_p$, when evidence $\{ e \}$ is provided by one of the argumentative policies. In Plot \ref{plt:argumentation_confuser} (green) we show the judge's accuracy when different argumentative agents propose $L=6$ required evidence for action $a_p$, averaged across $1000$ different debate games. The judge's accuracy is boosted from $65\%$ to a near $90\%$, demonstrating that agents can significantly amplify the capabilities of an otherwise limited judge.

\partitle{Robust Argumentation.}\label{sec:exp:argumentation:robustness}
A good supporting evidence is both convincing and not easily refutable by counterarguments. To test the robustness of the proposed evidence, we train $3$ adversarial \textit{confuser} agents \footnote{Confuser agents use the same architecture as argumentative agents, further described in App. \ref{appendix:models}.}, each targeting one of the three (frozen) argumentative agents. The goal of the confuser is to convince the judge of the opposing action $a_{1-p}$. For $L=6$, to obtain evidence $\{ e \}$, the agent (maxmin or self-play) and its corresponding confuser take turns and propose $3$ evidence each. The isolated agent is trained to first propose $L=3$ evidence, followed by the confuser proposing the remainder (see App. \ref{appendix:models:isolated} for more details and App. \ref{appendix:additional_results:isolated} for additional results). Plot \ref{plt:argumentation_confuser} (red) shows judge's accuracy in this setting for $1000$ different debate games. We observe that the isolated argumentative agent (Sec. \ref{sec:exp:environment:baselines}) is not resilient to refutations, enabling the confuser to bring the judge's accuracy down to $38\%$, effectively convincing it of the opposite of its preference. Differently, both maxmin and self-play agents managed to keep the judge's accuracy to about $85\%$.

\subsection{Experiment 4: Comparison to SHAP-based Explanations}\label{sec:exp:xai}
A widely used approach to analyze black-box machine learning models is through feature-attribution techniques. We aim to demonstrate that these explanations may not necessarily be as effective when used as supporting evidence. We focus on the SHAP framework \citep{lundberg2017unified}, specifically in providing justifications for decisions of various justifiable policies (Sec. \ref{sec:exp:trade_off}). To justify a decision using SHAP, we select the top $6$ state features, as ranked by their Shapely values. For argumentative models, we either run the debate between $a_0$ and $a_1$ (for maxmin and self-play agents) or propose $6$ evidence in isolation (for isolated agent). In Plot \ref{plt:shap}, we show the judge's accuracy against different approaches of proposing the evidence $\{ e \}$, across $1000$ comparisons $(s_t, a_0, a_1, p)$ sampled uniform-random from the test set. The SHAP-based evidence do improve the accuracy to about $70\%$, but nevertheless fall short compared to the argumentative agents.

\section{Discussion}
\label{sec:discussion}
In this work, we proposed use of a debate-based reward model as a general method of specifying desired behavior and necessary evidence to justify it. In this section, we take a step back and touch upon a couple of aspects that are relevant to future research based on this work.

\partitle{Eliciting Preferences.} The success of debate depends on the human's capability of judging its outcome, a process which may be affected by one's beliefs and biases. Extra care must be taken when collecting large datasets of preferences, as \textit{belief bias} \footnote{A \textit{belief bias} represents a tendency to judge the strength of arguments based on the plausibility of the conclusion, instead based on how strongly they support the conclusion.} is known to alter the results of judgment in human evaluations \citep{anderson2014belief} and is amplified in time-limited situations \citep{evans2005rapid}, which human annotators frequently encounter. Furthermore, in our experiments, we assumed existence of a single preference (clinician's decision). However, preferences collected from multiple raters will undoubtedly yield a higher variability in this respect. Motivated by positive results seen in this work, future research could undertake further studies that examine the effectiveness of eliciting preferences over partial state visibility through use of the debate as an amplification method.

\partitle{Defining Arguments is Difficult.}\label{sec:discussion:arguments} We have focused on debate assuming a well-defined argument space. While state features are one possible and easy choice, finding clear and expressive arguments presents a challenge, impacting the applicability of debate. In the context of RL, one potential alternative is considering previous trajectories in support of the current decision. In domains involving text generation, evidence could be defined as sentences, paragraphs, or references supporting a claim. A potentially interesting research direction is to examine the utility of a debate in a domain of human-ai collaboration, specifically in sequential decision-making tasks.

\partitle{Practical Considerations.}
We would like to emphasize that when deploying our framework in practice, it is important to account for the context in which the system is being employed. The performance-justifiability trade-off from our experiments suggests that a special care ought to be given to selecting hyperparameters, in particular, those that weight the importance of the environment rewards and debate-based rewards. In practice, this means that a reward designer has to assess the potentially differing objectives encoded in these values as well as the accuracy of the proxy judge model, prior to the training process. 

\section*{Ethics Statement} We acknowledge a potential misuse of the debate framework for malevolent purposes, such as deception. In the advent of AI systems that surpass human performance in many tasks, novel approaches must be developed which enable defense against malicious agents. While recent results indicate that debate is useful in thwarting malicious use of AI systems, further research is paramount in ensuring one can detect and defend against nefarious purposes. Moreover, the reliance on human judgments introduces the possibility of capturing their biases. Subsequently, designed reward function can incentivize argumentative agents to amplify those biases or learn to leverage them to achieve their objective. This is particularly important for practical considerations (Sec. \ref{sec:discussion}), where a reward designer is tasked with assessing and handling the performance-justifiability trade-off. It is important to note that our results do not hint at solutions for dealing with these challenges; they only demonstrate that the trade-off exists. It is therefore of great importance that future research investigates novel algorithmic approaches and methods to tackle such challenges.

\section*{Acknowledgments}
This research was, in part, funded by the Deutsche Forschungsgemeinschaft (DFG, German Research Foundation) -- project number $467367360$. We thank the anonymous reviewers for their valuable comments and suggestions.

\bibliography{iclr2024_conference}
\bibliographystyle{iclr2024_conference}

\appendix
\section{List of Appendices}
In this section, we provide a brief description of the content provided in the appendices of the paper.

\begin{itemize}
    \item Appendix \ref{appendix:dataset} provides details about the patient dataset and the defined environment:
        \begin{itemize}
            \item[o] \ref{appendix:dataset:sepsis} provides a general background on sepsis;
            \item[o] \ref{appendix:dataset:cohort} provides details about the patient cohort;
            \item[o] \ref{appendix:dataset:actions} provides details about the environment's action space;
            \item[o] \ref{appendix:dataset:rewards} provides details about the environment's reward structure.
        \end{itemize}
    \item Appendix \ref{appendix:models} provides details about the models:
        \begin{itemize}
            \item[o] \ref{appendix:models:judge} provides details about the judge model;
            \item[o] \ref{appendix:models:argumentation} provides details about the argumentative agents and their confusers;
            \item[o] \ref{appendix:models:sepsis} provides details about justifiable and baseline agents.
        \end{itemize} 
    \item Appendix \ref{appendix:additional_results} provides additional results:
        \begin{itemize}
            \item[o] \ref{appendix:additional_results:argumentation} provides results involving debates with $L=4$ evidence;
            \item[o] \ref{appendix:additional_results:preference_breakdown} provides further analysis of justifiable agent's actions;
            \item[o] \ref{appendix:additional_results:isolated} provides additional results pertaining to the isolated agent;
            \item[o] \ref{appendix:additional_results:rewdiff} provides results for agents using an alternative definition of the utility function;
            \item[o] \ref{appendix:additional_results:dataset} provides analysis of alternative methods for the preference dataset definition.
        \end{itemize}
\end{itemize}

\section{Dataset}\label{appendix:dataset}
\subsection{Sepsis}\label{appendix:dataset:sepsis}
Sepsis is a life-threatening condition, defined as severe infection leading to an acute organ dysfunction, which in turn can cause a cascade of changes that damage multiple organ systems \citep{singer2016third}. It is also one of the leading causes of patient mortality \citep{cohen2015}. Apart from administration of antibiotics and control of the infection source, a critical challenge in management of sepsis lies in the administration of intravenous fluids (IV) and vasopressors (VC). While international efforts attempt to provide a general guidance \citep{dellinger2004zl}, clinicians are nevertheless tasked in devising individualized treatments based on specificities of patients.

A first step towards automated management of septic patients was done in a seminal work of \citet{komorowski2018artificial}. The problem of treating sepsis was tackled by applying reinforcement learning to devise optimal treatment strategies for prescribing doses of intravenous fluids and vasopressors. The authors discretized the possible doses, which resulted in an action-space consisting of 25 distinct choices. The patient data was obtained from the MIMIC-III dataset \citep{johnson2016mimic}, and the authors extracted a subset of 48 patient features, discretized into 4h time windows. The preliminary results indicated that a learned policy was both quantitatively and qualitatively desirable. Since then, several other works extend and improve upon on this line of research. For example, \citet{raghu2017deep} proposed a continuous state-space approach based on deep reinforcement learning, on which we build on in this work. Likewise, \citet{huang2022reinforcement} proposed an approach that is based on continuous action-space, thus enabling more granular control of prescribed doses.

\subsection{Patient Cohort}\label{appendix:dataset:cohort}
To assemble our patient cohort, we utilize the MIMIC-III v1.4 database \citep {johnson2016mimic}, focusing our analysis on patients that fulfill the Sepsis-3 criteria \citep{singer2016third}. Similar as in \citet{komorowski2018artificial}, the sepsis is defined as a suspected existence of infection (indicated by the prescribed antibiotics) paired with a mild evidence of organ dysfunction (indicated by the SOFA score $\ge$ 2). To extract, preprocess and impute the data we utilize the pipeline described in \citet{komorowski2018artificial}, a process which resulted in 18,585 different patients that comprised our cohort. The cohort is split into three chunks of sizes 70\%, 15\%, 15\% representing the train, validation, and test splits. The observed mortality of the entire cohort was slightly above 6\%, and the splits were selected to approximately preserve this ratio. The course of a patient treatment is represented as a trajectory consisting of state-action pairs, terminating upon patient discharge or death. The average trajectory length was 13, with 2 being the smallest and 20 being the largest.  The state is a 44 dimensional vector, comprised of 40 time-varying continuous and 4 demographic discrete features, shown in Table \ref{table:features}. Patients’ data were discretized using 4h time steps, where variables with multiple measurements within this window were averaged (e.g., heart rate) or summed (e.g., urine output) as appropriate.
\begin{table}[t]
    \caption{List of features comprising the patient vector. We select 40 time-varying continuous features and 4 demographic discrete features.}
    \label{table:features}
    \begin{center}
        \renewcommand{\arraystretch}{1.3}
        \begin{tabular}{ |c|c|c| }
            \multicolumn{3}{l}{\underline{Continuous features}} \\
            \hline
            SOFA & Calcium & Shock index \\ 
            Urine output 4h & Urine output total & Cumulated balance \\
            Glasgow coma scale & Heart rate & Systolic blood pressure \\
            Mean blood pressure & Diastolic blood pressure & Total input fluids  \\
            Respiratory rate & Temperature & FiO2 - fraction of inspired oxygen \\
            Potassium & Sodium & Chloride \\
            Glucose & Magnesium & SIRS \\
            Hemoglobin & White blood cells count & Platelets count \\
            Partial Thromboplastin Time & PH - Acidity & PaO2 \\
            PaCO2 & Base excess & Bicarbonate \\
            Lactate & PaO2/FiO2 Ratio & Oxygen saturation \\
            BUN & Creatinine & SGOT \\
            SGPT & Total bilirubin & International normalized ratio \\
            Prothrombin time & & \\
            \hline
            \multicolumn{3}{l}{\underline{Discrete features}} \\
            \hline
            Age & Gender & Weight \\
            Mechanical ventilation & & \\
            \hline
        \end{tabular}
    \end{center}
\end{table}

\subsection{Action Space}\label{appendix:dataset:actions}
\partitle{Argumentative Policies.} The evidence (action) space of the argumentative policies is defined by the total number of state features, $44$ in our case, as listed in Table \ref{table:features}. To prevent the agent from repeating already proposed arguments, we additionally employ \textit{action masking}, setting the log probability of already presented arguments to negative infinity.

\partitle{Baseline and Justifiable Policy.} To devise an action space of policies treating sepsis, we follow previous work \citep{komorowski2018artificial,raghu2017deep} and focus on managing the total volume of intravenous fluids (IV) and maximum dose of vasopressors (VC) administered to the patient over a $4\text{h}$ discretization window. The dose of each treatment is represented as one of four possible non-null choices derived from observed doses divided into four quartiles. We additionally define another choice, designated as an option 'no drug given'. The combination of these produced $25$ possible discrete actions, $5$ per each treatment, comprising the action space of the policy.

\subsection{Rewards}\label{appendix:dataset:rewards}

\partitle{Sepsis.} To define intermediate and terminal rewards for treating septic patients, following the work of \citet{komorowski2018artificial}, we defined the primary outcome of the treatment via 90-day patient mortality. Therefore, the agent's objective is to optimize for patient survival. To this end, we issue terminal environment rewards of $\pm 15$ for every patient discharge and death, respectively. To stabilize the training of a deep-rl policy, we also issue intermediate rewards that are clinically guided, as in \citet{raghu2017deep}. These rewards are comprised of fairly reliable indicators of the patient's overall health, namely the SOFA score (measure of organ failure) and a patient’s lactate levels (measure of cell-hypoxia, which is usually higher in septic patients). The rewards for intermediate time steps are then shaped as follows:
\begin{equation*}
    r(s_t, s_{t+1}) = C_0 \mathbbm{1}(s_{t+1}^{\text{SOFA}} = s_t^{\text{SOFA}} \& s_{t+1}^{\text{SOFA}} > 0) + C_2 (s_{t+1}^{\text{SOFA}} - s_t^{\text{SOFA}}) + C_2 \text{tanh}(s_{t+1}^{\text{Lactate}} - s_t^{\text{Lactate}})
\end{equation*}
where $C_0$, $C_1$ and $C_2$ are tunable parameters which we set to $C_0 = -0.025$, $C_1 = -0.125$ and $C_2 = -2$, following previous work of \citet{raghu2017deep}. Rewards defined in this way penalize both, the high SOFA scores and lactate values, as well as increases in these quantities.

\section{Models}\label{appendix:models}

\subsection{Judge}\label{appendix:models:judge}
We defined a judge as a scalar reward function $\mathcal{J}_\theta (a, \{e\}) \in \mathbb{R}$ that quantifies the level of support a decision $a$ has by the set of evidences $\{e\}$. The judge is parametrized by weights $\theta \in \mathbf{R}^{d_1}$ of a neural network with two hidden layers of size $256$, using parametric relu \citep{he2015delving} activation and batch normalization \citep{ioffe2015batch}. The network takes as input values of proposed evidence $\{e\}$, a binary mask wherein all elements corresponding to the evidence $\{e\}$ are assigned a value of one, while the remaining elements are set to zero, as well as a one-hot encoded action. The addition of a binary mask empirically led to a more stable learning. During training, we augment a tuple $(s_t, a_0, a_1, p) \sim \mathcal{D}$ with an evidence set of state features sampled uniform-random from the state $s_t$ i.e., $\{e\} \sim \mathcal{U}$, anew for each training batch. To learn the parameter $\theta$, we minimize the cross-entropy loss between preference-predictions and labels from the dataset:
\begin{equation*}
\min_{\theta \in \mathbb{R}^d} - \mathbb{E}_{(s_t, \{ e \}, a_0, a_1, p) \sim \mathcal{D}} \left[ 
p \cdot \log{\mathcal{P}(a_1 \succ a_0, \{ e \})} + (1 - p) \cdot \log{\mathcal{P} (a_0 \succ a_1, \{ e \})} \right]
\end{equation*}
The learning is done for a total of $100$ epochs using a batch size of $64$, Adam optimizer and a learning rate of $5\text{e-}4$.

\subsection{Argumentation}\label{appendix:models:argumentation}
All argumentative models utilize the same network architecture comprised of $2$ hidden layers of size $512$ with a leaky-relu activation function with slope of $1\text{e-}2$. The network input consists of a $44$ dimensional patient state vector, the decision for which the agent is arguing, as well as a binary mask of arguments proposed thus far. The action space of the agent is represented by all $44$ state features (see App. \ref{appendix:dataset:actions} for more details). For training, we use PPO \citep{schulman2017proximal}, running the procedure for $1$M steps using the Adam optimizer with a learning rate $5\text{e-}4$ and a batch size of $128$. The discount factor was empirically tuned and set to $0.9$. The full list of hyperparameters and their considered tuning ranges is given in Table \ref{table:argumentation:hp}. During training, the agent's policy is stochastic: the evidence is sampled from a categorical distribution defined by its logits. To obtain a deterministic policy used in evaluations, we perform an \textit{argmax} operator over obtained logits in a particular state.

\subsubsection{Isolated Agent}\label{appendix:models:isolated}
When investigating robustness of the multi-agent debate, we examine two different setups involving an isolated agent baseline: \textit{precommit} (reported in the main paper, Plot \ref{plt:argumentation_confuser}) and \textit{adaptive} (reported in the App. \ref{appendix:additional_results:isolated}, Plot \ref{plt:appendix:isolated_agent}). The former represents an easier case for the agent, but lacks the debate structure. The latter uses a full debate setup as described in Sec. \ref{sec:formal:debate}, but evaluates the isolated agent in a setup slightly different from one it was trained in.

\partitle{Precommit.} In this case, an isolated agent is trained to propose evidence $\{e\}$ of size $L/2$ that for a given $a_p$ maximizes $\mathcal{J}^{\prime}(a_p, \{e\})$, where $\mathcal{J}^{\prime}$ is a new judge trained to evaluate $L/2$ evidence. When evaluating robustness (Plot \ref{plt:argumentation_confuser}), the agent first proposes $L/2$ evidence, followed by a confuser agent which proposes the remainder.

\partitle{Adaptive.} In this case, an isolated agent is trained to propose all $L$ evidence $\{e\}$ that maximize $\mathcal{J}(a_p, \{e\})$, for a given $a_p$. When evaluating robustness (Plot \ref{plt:appendix:isolated_agent}), the isolated agent and its associated confuser take turns and propose a total of $L/2$ evidence each.

\subsubsection{Debate Agents}
For multi-agent scenarios, both \textit{maxmin} and \textit{self-play} agents use the architecture described in the beginning of Sec. \ref{appendix:models:argumentation}, but each modify the underlying optimization pipeline. To train a \textit{self-play} debate agent, we let the agent argue with a (frozen) copy of itself, updating it every $100$k steps. This procedure is then repeated for a fixed number of $500$ generations. To train the \textit{maxmin} debate agent, we parametrize the agent's opponent with a different set of weights $\phi_2 \in \mathbb{R}^{d_2}$. The procedure starts by training the main agent for $4$k steps, followed by training of its opponent for $100$k steps. Like in the previous case, the procedure is repeated for $500$ generations. This approach is based on the bi-level optimization and allows for overfitting the opponent to the current version of the main agent, which ensures learning of defense strategies against a very strong adversary.

\subsubsection{Confuser Agents}
The evaluation of robustness we presented in Section \ref{sec:exp:argumentation:robustness} required learning three separate adversaries (one for each argumentative agent), explicitly tasked in providing counterarguments that will confuse the judge. The architecture of these \textit{confuser} agents is mostly the same as that of the argumentative agents, shown in Table \ref{table:argumentation:hp}. For a sample $(s_t, a_0, a_1, p) \sim \mathcal{D}$, the confuser agent is rewarded positively, whenever the judge is convinced of the alternative (non-preferred) action.

\begin{table}[t]
    \caption{Hyperparameters used for argumentative agents. Unless otherwise indicated, all agents utilize the same parameters.}
    \label{table:argumentation:hp}
    \begin{center}
        \renewcommand{\arraystretch}{1.3}
        \begin{tabular}{ |c|c|c| }
            \multicolumn{3}{l}{\underline{Common parameters}} \\
            \hline
            \textbf{Parameter name} & \textbf{Parameter value} & \textbf{Tuning range} \\
            \hline
            Hidden dim & 512 & [256, 512] \\
            Learning rate & 5e-4 & loguniform[1e-5:1] \\
            Entropy coefficient & 1e-2 & loguniform[0.00000001:0.1] \\ 
            Clip range & 0.1 & [0.1, 0.2, 0.3, 0.4] \\
            Discount & 0.9 & [0.8, 0.9, 0.95, 0.99] \\
            GAE lambda & 0.7 & [0.7, 0.8, 0.9, 0.92, 0.95, 0.98, 0.99, 1.0] \\
            VF weight & 0.5 & [0.3, 0.5, 0.65, 0.75] \\
            Max grad norm & 0.1 & [0.3, 0.5, 0.6, 0.7, 0.8, 0.9, 1, 2, 5] \\
            Normalize rewards & true & [true, false] \\
            Ortho init & true & [true, false] \\
            \hline
            \multicolumn{3}{l}{\underline{Confuser parameters}} \\
            \hline
            Hidden dim & 256 & [256, 512] \\
            Entropy coefficient & 3e-4 & loguniform[0.00000001:0.1] \\
            Clip range & 0.4 & [0.1, 0.2, 0.3, 0.4] \\
            VF weight & 0.65 & [0.3, 0.5, 0.65, 0.75] \\
            Max grad norm & 2 & [0.3, 0.5, 0.6, 0.7, 0.8, 0.9, 1, 2, 5] \\
            \hline
        \end{tabular}
    \end{center}
\end{table}

\subsection{Baseline and Justifiable Agents}\label{appendix:models:sepsis}
Our approach for learning treatment policies for septic patients is based on continuous state-space models and builds on \citet{raghu2017deep}. The policy is based on a variant of Deep-Q networks \citep{mnih2015human}, specifically double-deep \citep{van2016deep} Q-network, also employing the dueling architecture \citep{wang2016dueling}, where the estimated action-value function for a pair $(s, a)$ is split into separate \textit{value} and \textit{advantage} streams. The final network consists of $2$ fully-connected layers of size $128$ using leaky-relu activation functions with slope $1\text{e-}2$. The learning is done in batches of $256$ $(s, a, r, s')$ tuples sampled from a Prioritized experience replay buffer \citep{schaul2015prioritized} with a learning rate of $1\text{e-}4$. Instead of periodically updating the target network, we leverage Polyak averaging with an update coefficient $\tau$ set to $1\text{e-}3$. The full list of used hyperparameters is given in Table \ref{table:protagonist:hp}.
\begin{table}
    \caption{Hyperparameters used for baseline and justifiable policies. Unless otherwise indicated, all policies use the same parameters. If a parameter does not specify a tuning range, its value has either been selected based on previous work (e.g., buffer $\alpha$ and $\beta$) or tuning was not necessary (e.g., debate coefficient $\lambda$).}
    \label{table:protagonist:hp}
    \begin{center}
        \renewcommand{\arraystretch}{1.3}
        \begin{tabular}{ |c|c|c| }
            \multicolumn{2}{l}{\underline{Baseline policy}} \\
            \hline
            \textbf{Parameter name} & \textbf{Parameter value} & \textbf{Tuning Range} \\
            \hline
            Hidden dim & 128 & [64, 128, 256] \\
            Learning rate & 1e-4 & loguniform[1e-5:1e-1] \\
            Batch size & 256 & [128, 256, 512, 1024] \\
            Polyak update & 1e-3 & [1e-1, 1e-2, 1e-3, 1e-4] \\
            ReLu slope & 1e-2 & [1e-1, 1e-2, 1e-3] \\
            Discount & 0.99 & - \\
            Num. estimation step & 6 & [1, 3, 6, 10, 15] \\
            Terminal reward & $\pm$ 15 & - \\
            Debate scaling coefficient $\alpha$ & $\pm$ 5 & - \\
            Debate coefficient $\lambda$ & 0.0 & - \\
            Buffer $\alpha$ & 0.6 & - \\
            Buffer $\beta$ & 0.9 & - \\
            \hline
            \multicolumn{3}{l}{\underline{$\lambda \in$ [0.25, 0.5, 0.75]}} \\
            \hline
            Num. estimation step & 3 & - \\
            \hline
            \multicolumn{3}{l}{\underline{$\lambda = 1.0$}} \\
            \hline
            Num. estimation step & 1 & - \\
            \hline
        \end{tabular}
    \end{center}
\end{table}
Similar to \citet{raghu2017deep}, we augment the standard Q-network loss. First, we added a regularization term that penalizes Q-values outside the allowed threshold $Q_{\text{thresh}} = \pm 20$. In addition, we clip the target network outputs to $\pm 20$, which empirically proved to stabilize the learning. The final loss function we used is given by:
\begin{align*}
    \mathcal{L}(\Theta) =& \mathbb{E} \left[(Q_{\text{double-target}} - Q(s, a; \Theta))^2\right] + \beta \cdot \text{max}(|Q(s, a; \Theta)| - Q_{\text{thresh}}, 0) \\
    Q_{\text{double-target}} =& r + \gamma \cdot Q(s^{\prime}, \text{arg max}_{a^{\prime}} Q(s^{\prime}, a^{\prime}; \Theta); \Theta^{\prime})
\end{align*}
Where $\beta$ is a user-specified coefficient we set to $\beta=5.0$ in all our experiments following \citet{raghu2017deep}. The learning is done for a total of $25$k iterations, evaluating the policy every $50$ iterations using weighted importance sampling (WIS) \citep{sutton2018reinforcement} on a held-out test set. 

\partitle{Behavioral Policy.} The calculation of the importance sampling ratio requires access to a so-called \textit{behavioral policy} that generated offline samples. To obtain it, we train a behavior-cloning (BC) clinician policy to take in a state $s_t$ from the patient cohort and predict the action $a_t$ taken by the human clinician, minimizing the cross-entropy loss over the training dataset. The network consists of two fully-connected layers of size $64$. We run the training with the Adam optimizer, using a learning rate of $1\text{e-}3$ and a weight decay set to $1\text{e-}1$ for a total of $100$ epochs with a batch size of $64$.

\section{Additional Results}\label{appendix:additional_results}

\subsection{Shorter Debates}\label{appendix:additional_results:argumentation}
In the main text, we have seen positive results involving debate using $6$ arguments, which amounts to $\sim$ 13\% of the entire state. In this section, we want to further examine the effectiveness of debate when limiting the number of evidence to $L=4$, amounting to 9\% of the entire state.

\begin{figure*}
    \centering
    \begin{subfigure}[b]{0.24\textwidth}
        \centering
        \includegraphics[width=\textwidth]{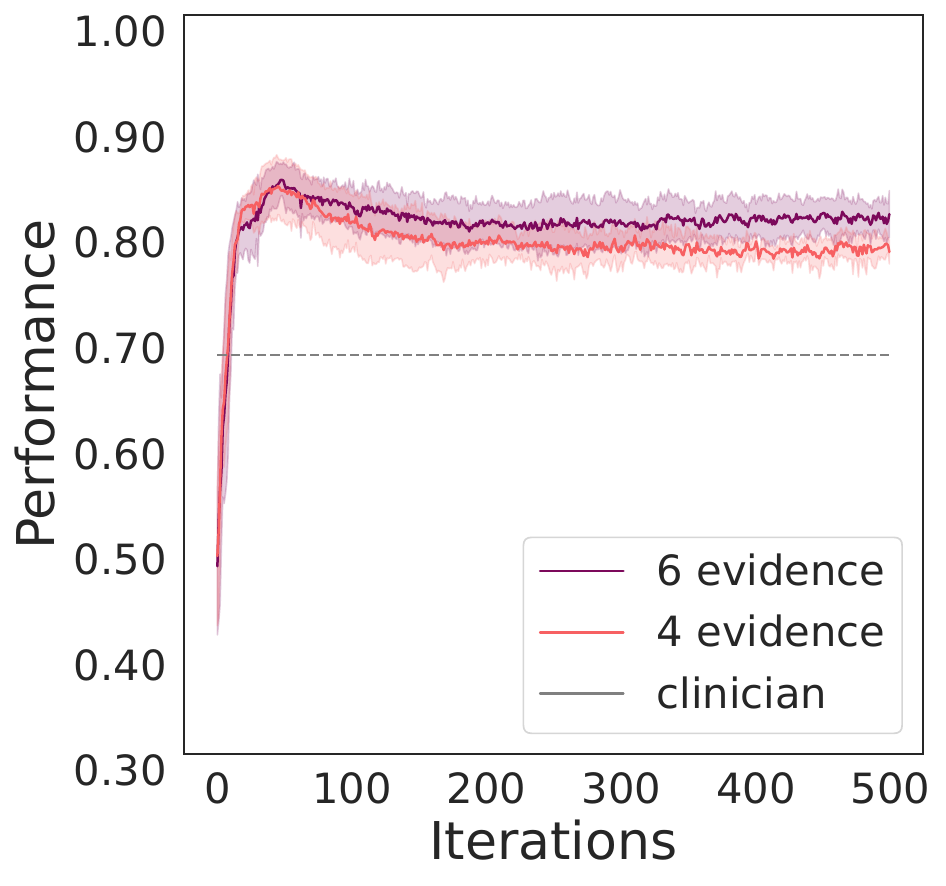}
        \caption{$\lambda=0.25$}
        \label{plt:appendix:wis_l25}
    \end{subfigure}
    \hfill
    \begin{subfigure}[b]{0.24\textwidth}
        \centering
        \includegraphics[width=\textwidth]{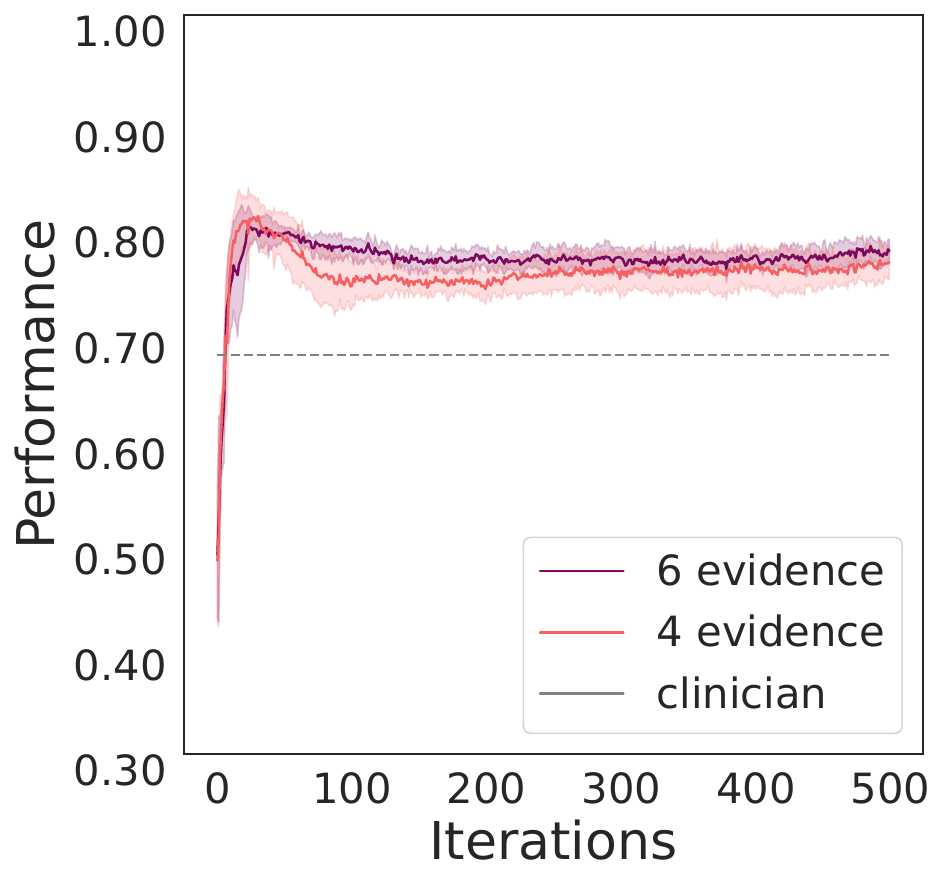}
        \caption{$\lambda=0.5$}
        \label{plt:appendix:wis_l50}
    \end{subfigure}
    \begin{subfigure}[b]{0.24\textwidth}
        \centering
        \includegraphics[width=\textwidth]{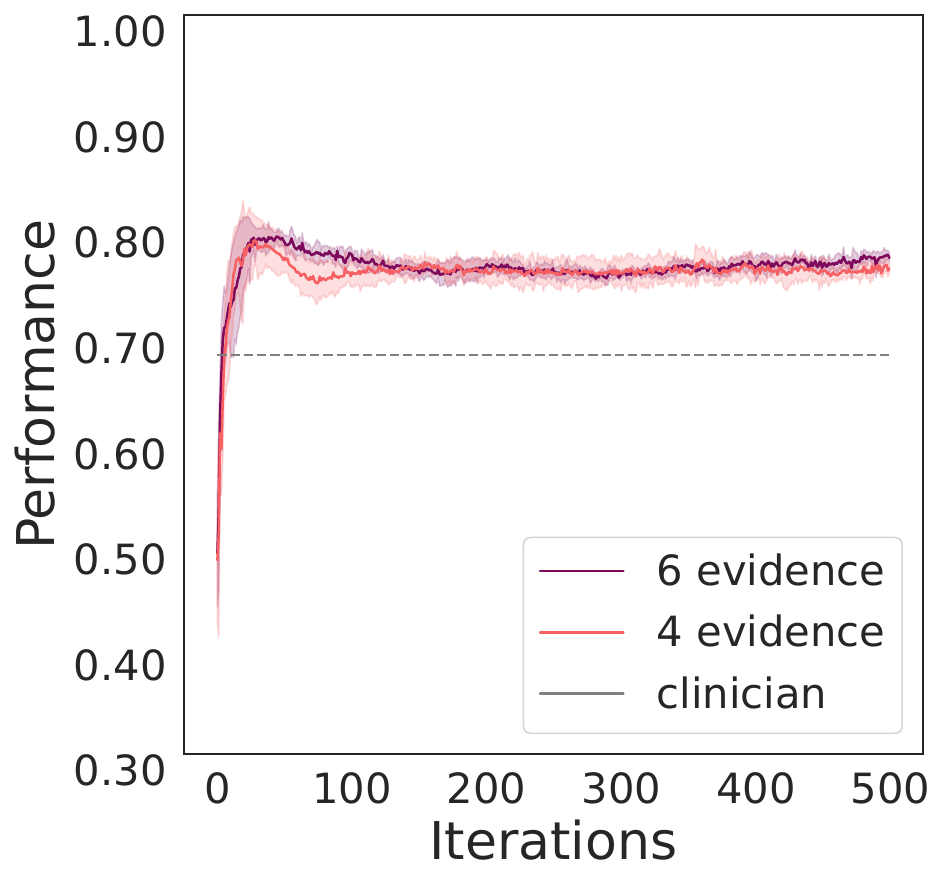}
        \caption{$\lambda=0.75$}
        \label{plt:appendix:wis_l75}
    \end{subfigure}
    \hfill
    \begin{subfigure}[b]{0.24\textwidth}
        \centering
        \includegraphics[width=\textwidth]{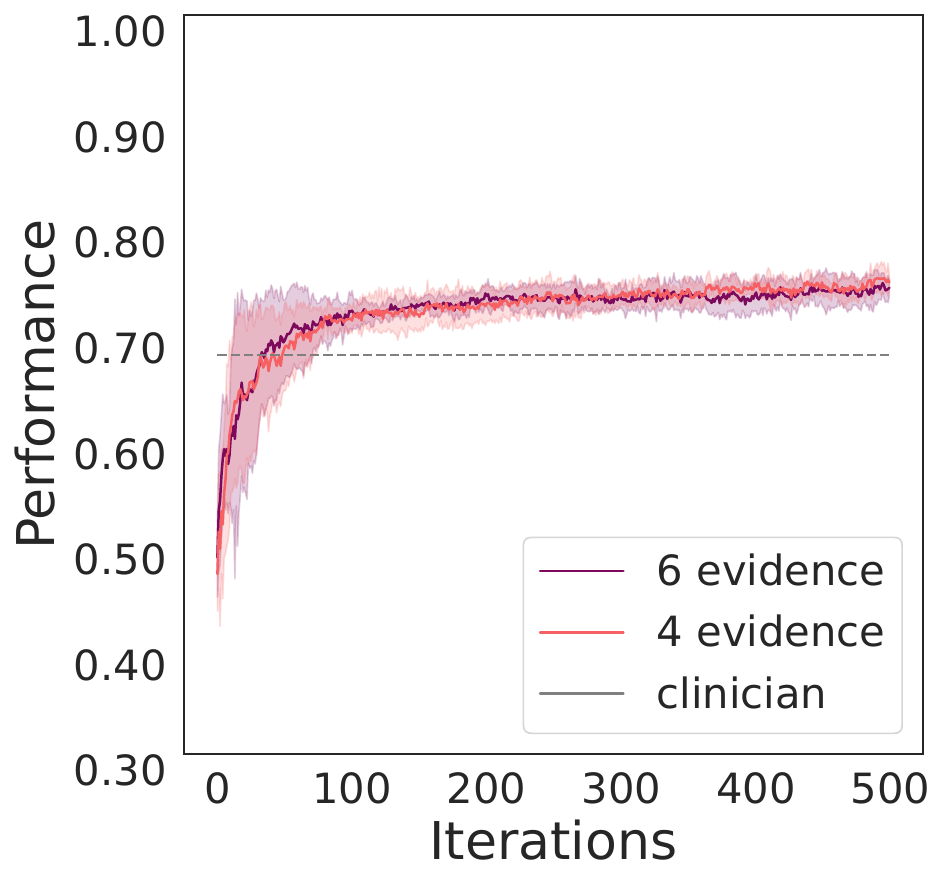}
        \caption{$\lambda=1.0$}
        \label{plt:appendix:wis_l100}
    \end{subfigure}
    
    \begin{subfigure}[b]{0.24\textwidth}
        \centering
        \includegraphics[width=\textwidth]{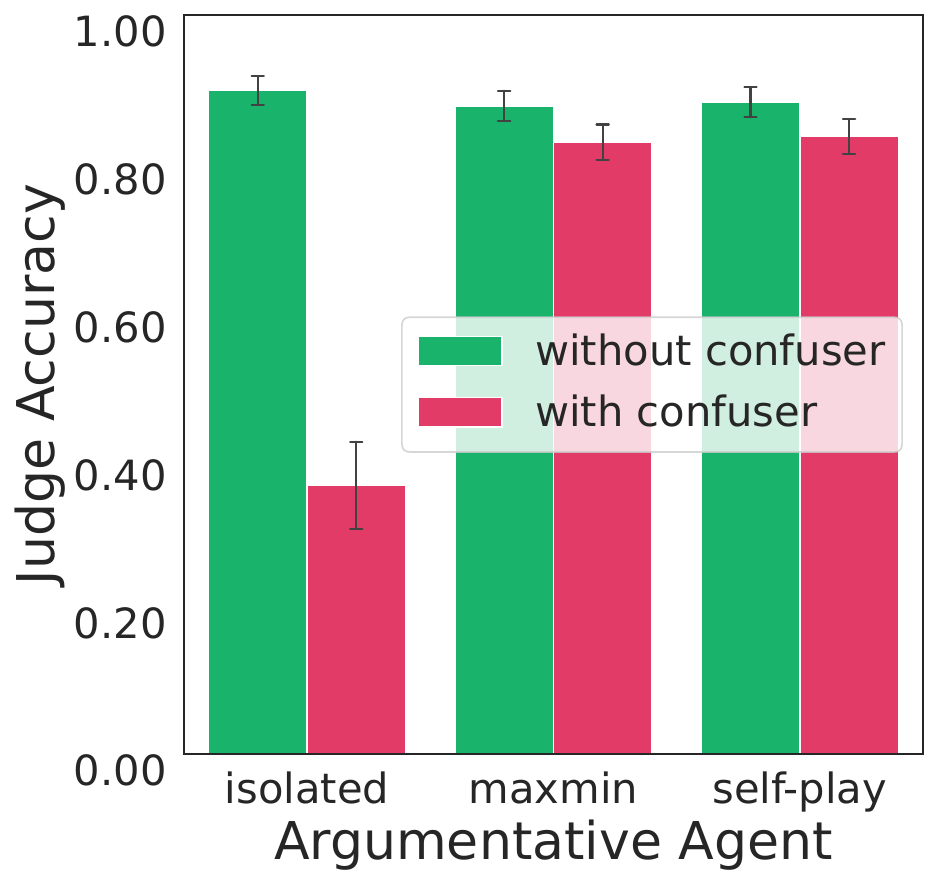}
        \caption{Convincing success}
        \label{plt:appendix:preference_recovery}
    \end{subfigure}
    \hfill
    \begin{subfigure}[b]{0.24\textwidth}
        \centering
        \includegraphics[width=\textwidth]{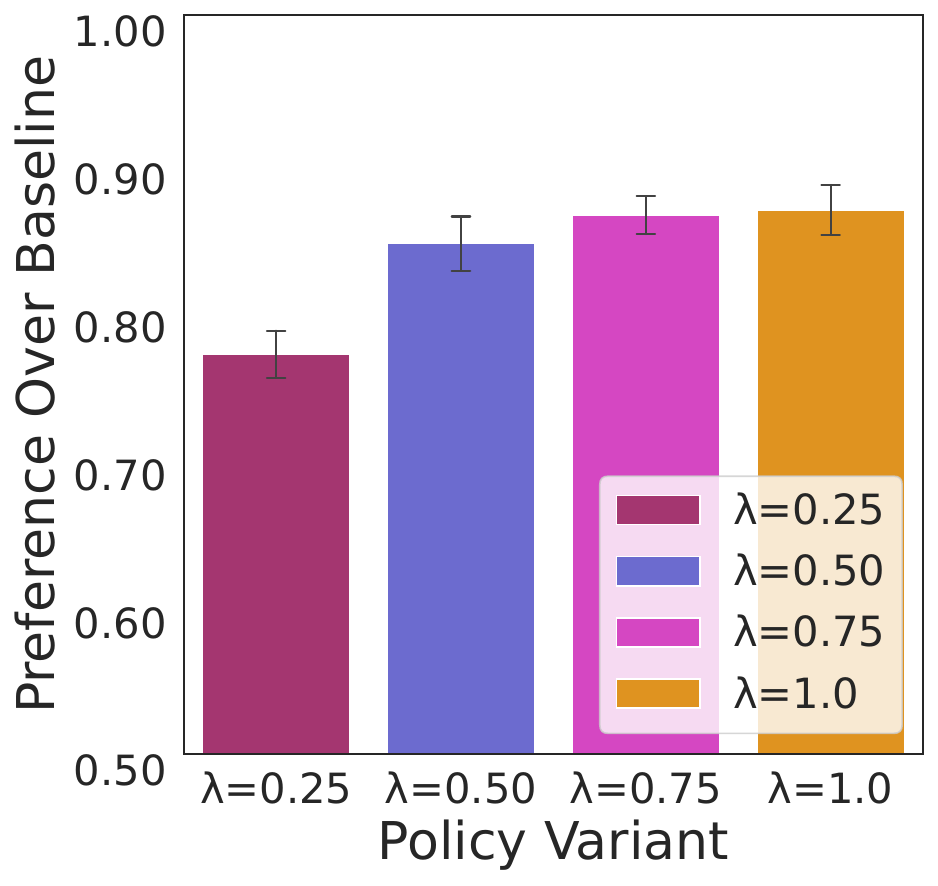}
        \caption{Preferred to baseline}
        \label{plt:appendix:jstf}
    \end{subfigure}
    \begin{subfigure}[b]{0.24\textwidth}
        \centering
        \includegraphics[width=\textwidth]{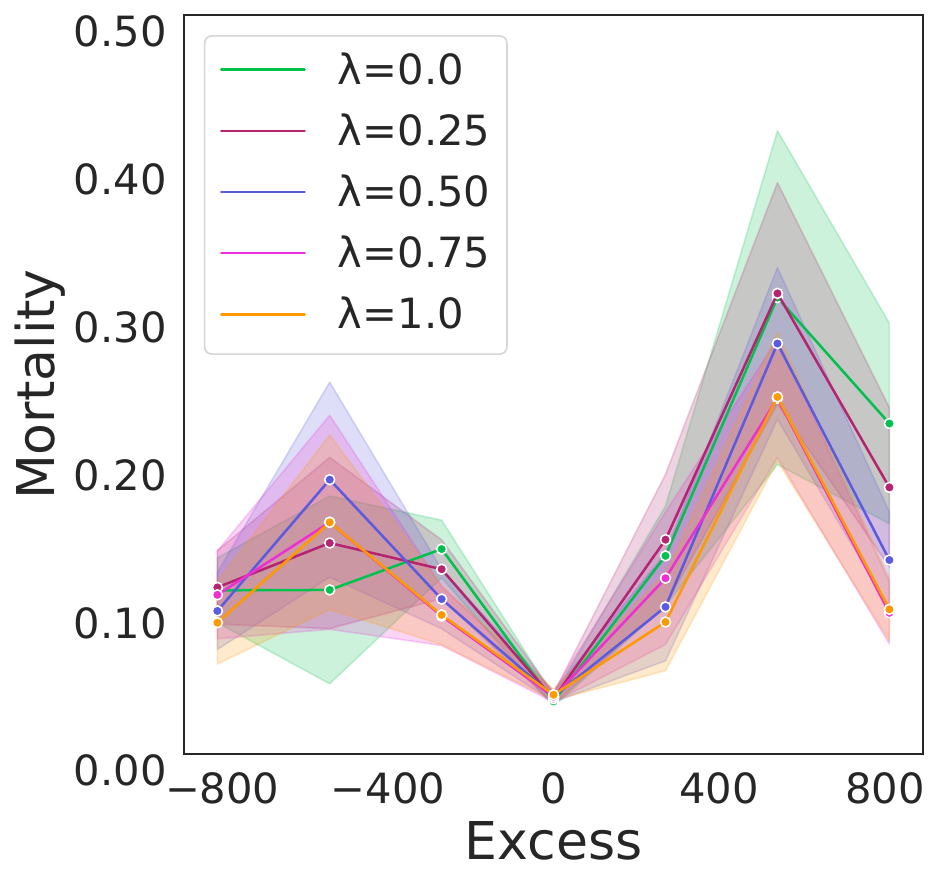}
        \caption{IV dose excess}
        \label{plt:appendix:iv}
    \end{subfigure}
    \hfill
    \begin{subfigure}[b]{0.24\textwidth}
        \centering
        \includegraphics[width=\textwidth]{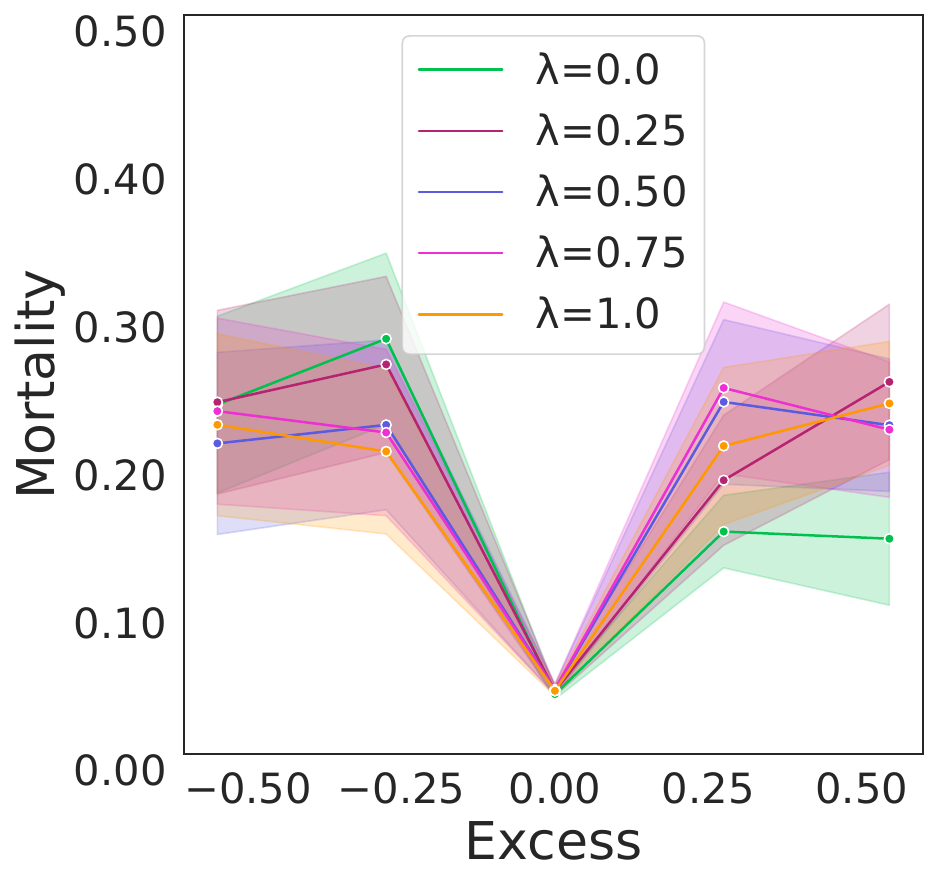}
        \caption{VC dose excess}
        \label{plt:appendix:vc}
    \end{subfigure}
    \hfill
    \caption{Quantitative and qualitative evaluation of policies trained using debate-based rewards limited to $L=4$ evidence. For (f)-(h), the confidence intervals represent $\pm 2$ standard errors of the mean over $5$ random seeds. (a)-(d) Performance of policies trained with $L=4$ and $L=6$ evidence, as measured by WIS evaluation on a held-out test set with $\pm1$ terminal rewards for every patient discharge or death. The mean and standard deviation are reported over $5$ random seeds. (e) Accuracy of the judge in predicting the preferred action using $4$ proposed evidence, with and without the confuser agent. The CI represent $\pm 2$ standard errors of the mean estimate. (f)  Percent of times judge preferred decisions of justifiable policies (i.e., $\lambda > 0.0$) compared to those of the baseline policy (i.e., $\lambda = 0.0$). (g) (h)  Observed patient mortality (y-axis) against variations in IV/VC treatment doses prescribed by clinicians compared to the recommendations of learned policies (x-axis).}
    \label{fig:protagonist_evaluation_wis_4arg}
\end{figure*}

The first question that arises when further limiting the amount of state visibility during debate is the impact it has on the quality of learned task policies. In Plots 
\ref{plt:appendix:wis_l25}-\ref{plt:appendix:wis_l100}, we evaluate the performance of different justifiable policies on a held-out test set during the course of training using weighted importance sampling, for the case of $4$ and $6$ evidence limit. While the reduced number of evidence exposes the judge to only 9\% of the entire state, we can see that the achieved performance is comparable to the case where $L=6$. Furthermore, in Plot \ref{plt:appendix:jstf}, we also confirm that trained policies are significantly more preferred to the baseline policy, a trend equivalent to the one we saw in the main text. Apart from these quantitative evaluations, in Plots \ref{plt:appendix:iv} and \ref{plt:appendix:vc}, we show the qualitative analysis of patient mortality from Sec. \ref{sec:exp:trade_off}. We confirm that the lowest mortality is observed when clinician prescribed doses recommended by justifiable policies, thus further signaling potential validity of policies trained with rewards stemming from debates with $L=4$ evidence.

To evaluate the effectiveness of argumentative agents, we repeat the evaluation from Sec. \ref{sec:exp:argumentation:prr}. When exposed to $4$ randomly selected evidence, the judge trained via the procedure outlined in App. \ref{appendix:models:judge} achieves accuracy of $\sim$59\%. Plot \ref{plt:appendix:preference_recovery} shows the judge's accuracy when different argumentative agents propose $L=4$ evidence \footnote{Like in the main text, we use the \textit{precommit} setup for the isolated agent (see App. \ref{appendix:models:isolated} for more details).}, averaged over $1000$ samples $(s_t, a_1, a_2, p) \sim \mathcal{D}$. Without the confuser, all three agents achieve performance similar to the case of $L=6$ evidence, boosting the judge's accuracy to almost $90\%$. In the setup involving an adversary, agents propose a total of $2$ evidence each before the judge evaluates the outcome. The observed trend is also similar to the one from the main text (Sec. \ref{sec:exp:argumentation}).

\begin{figure*}
    \centering
    \hfill
    \begin{subfigure}[b]{0.24\textwidth}
        \centering
        \includegraphics[width=\textwidth]{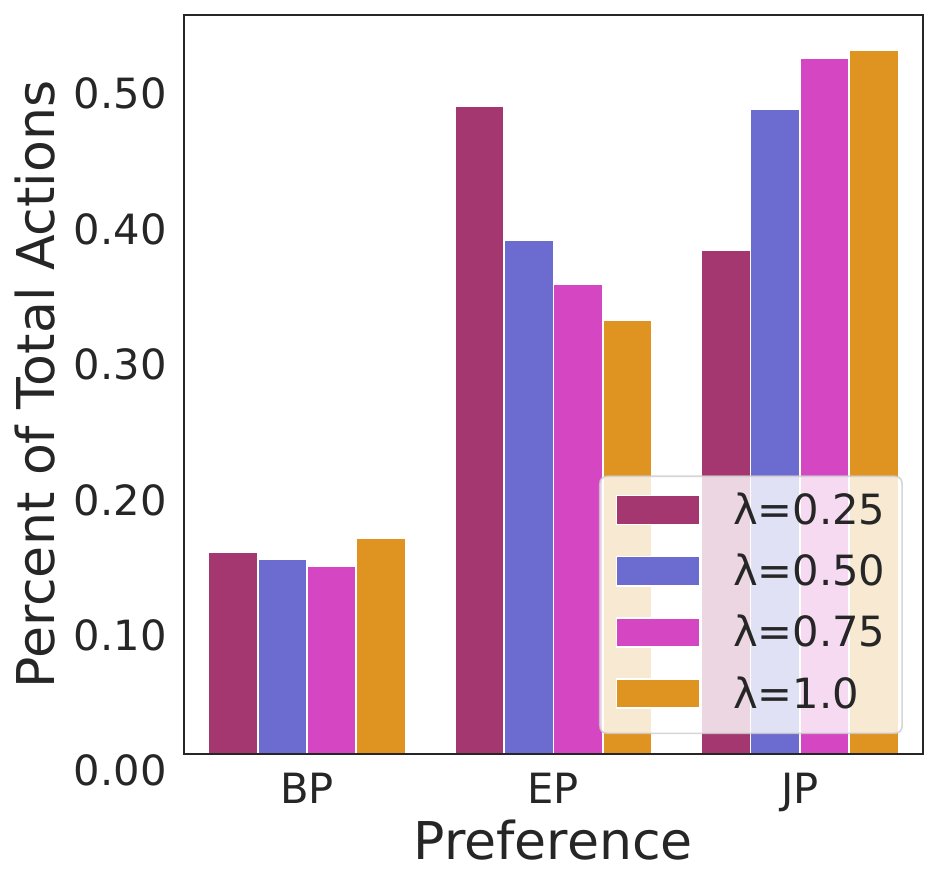}
        \caption{Preference breakdown}
        \label{plt:appendix:preference_breakdown}
    \end{subfigure}
    \begin{subfigure}[b]{0.24\textwidth}
        \centering
        \includegraphics[width=\textwidth]{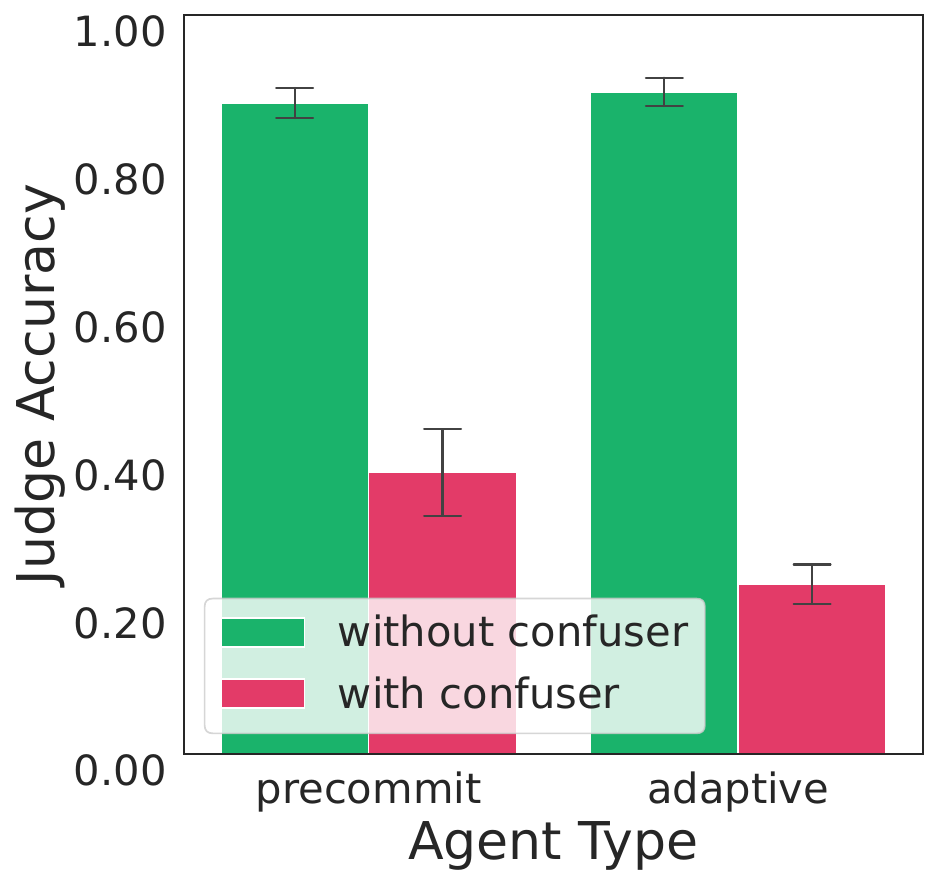}
        \caption{Isolated agent setup}
        \label{plt:appendix:isolated_agent}
    \end{subfigure}
    \begin{subfigure}[b]{0.243\textwidth}
        \centering
        \includegraphics[width=\textwidth]{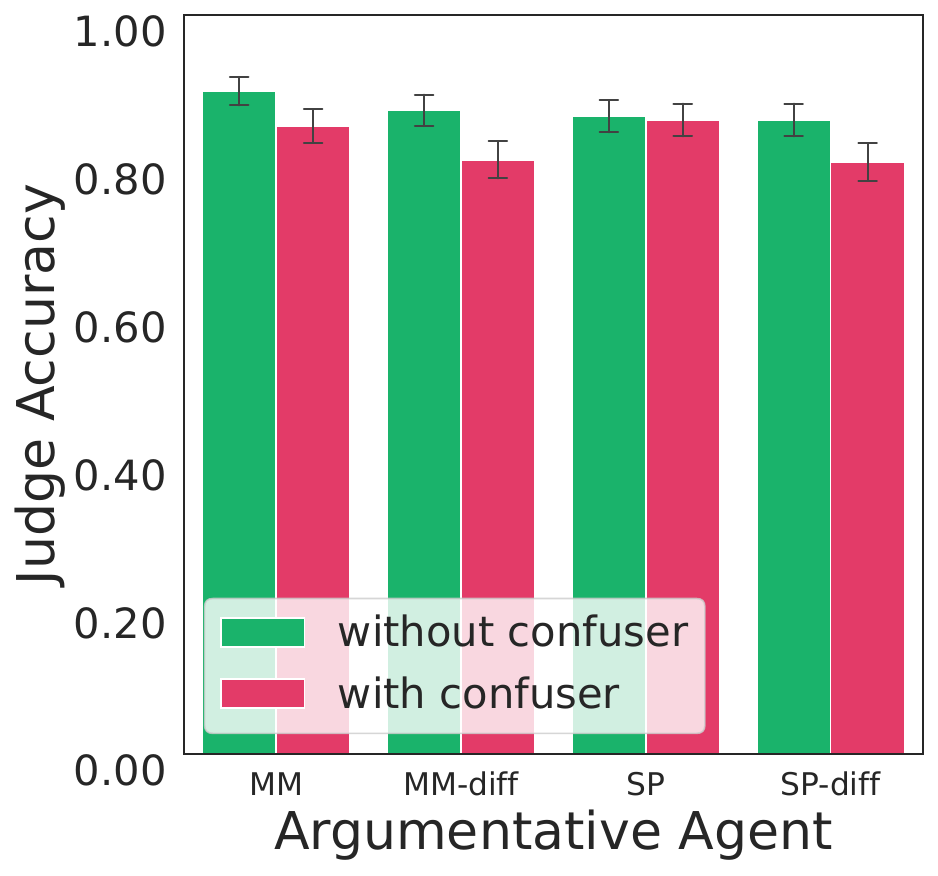}
        \caption{Convincing success}
        \label{plt:appendix:argumentation_rewdiff}
    \end{subfigure}
    \hfill
    \begin{subfigure}[b]{0.24\textwidth}
        \centering
        \includegraphics[width=\textwidth]{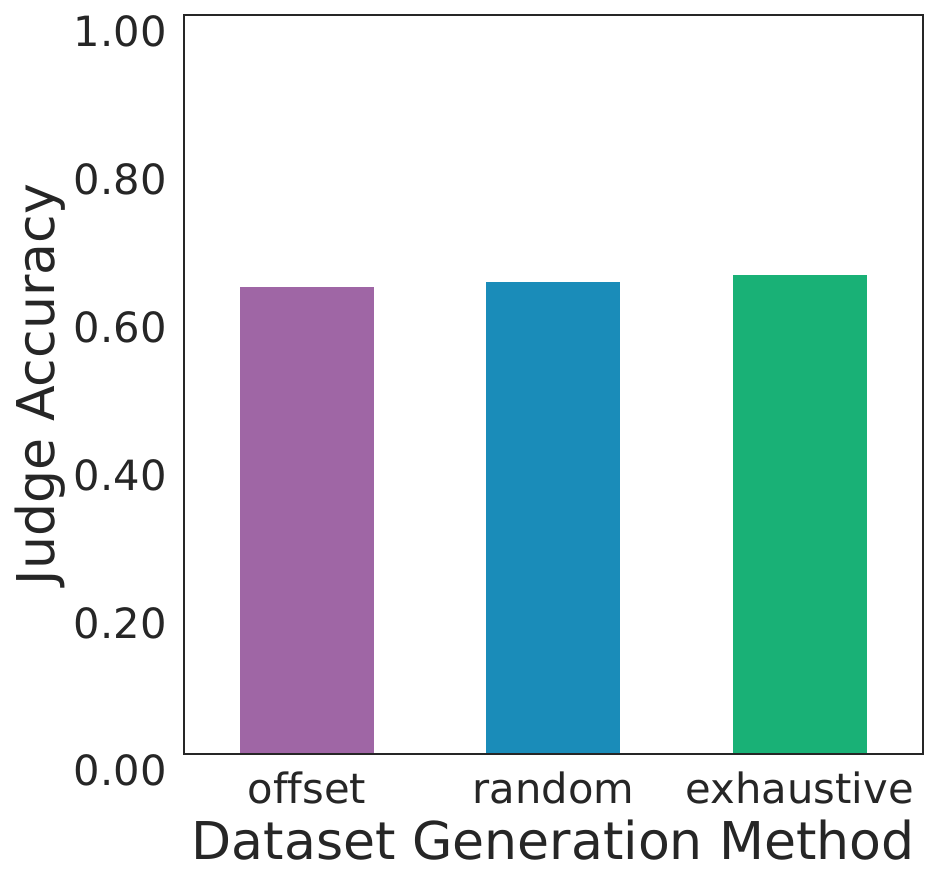}
        \caption{Preference datasets}
        \label{plt:appendix:dataset_generation_method}
    \end{subfigure}
    \hfill
    \caption{(a) Percent of times when actions of the justifiable policy were more preferred to those of the baseline policy (JP), less preferred (BP) and equally preferred (EP). (b) Accuracy of the judge in predicting the preferred action, with and without the confuser agent when evidence was proposed by two isolated agents trained in \textit{precommit} and \textit{adaptive} setup. The CI represents $\pm2$ standard errors of the mean estimate. (c) Accuracy of the judge in predicting the preferred action, when evidence was proposed by maxmin (MM) and self-play (SP) agents trained using the utility function proposed in Sec. \ref{sec:formal:debate} compared to agents trained using the utility function defined as a difference in predicted rewards proposed in App \ref{appendix:additional_results:rewdiff}, MM-diff and SP-diff respectively. The CI represents $\pm2$ standard errors of the mean estimate. (d) Accuracy of the judge in predicting the preferred action with evidence sampled uniform-random, trained on datasets generated with different methods.}
\end{figure*}

\subsection{Preference Breakdown}\label{appendix:additional_results:preference_breakdown}
To gain a better understanding about the behavior of learned justifiable policies, we further examine actions they propose. In particular, Plot \ref{plt:appendix:preference_breakdown} shows the percent of times an action from the justifiable policy was preferred to that of the baseline policy (JP), percent of times when the action of the baseline policy was preferred to the action of the justifiable policy (BP), and percent of time when the two were equally preferred (EP) \footnote{In our experiments, the judge deemed two actions equally justifiable whenever they were the same.}. We see that the number of actions proposed by the justifiable agent and the baseline agent that are equally preferred decreases as the parameter $\lambda$ is increased: for $\lambda=0.25$ two agents chose the equally justifiable action about $49\%$ of the times, for $\lambda=0.50$ this number drops to $40\%$, whereas for $\lambda=0.75$ and $\lambda=1.0$ this number is $36\%$ and $35\%$ respectively. Out of the remaining actions, the ones from justifiable policies were increasingly more preferred, as the parameter $\lambda$ was increased.

\subsection{Isolated Agent Setup}\label{appendix:additional_results:isolated}
In App. \ref{appendix:models:isolated}, we described two setups which we consider when evaluating robustness of the isolated agent, namely \textit{precommit} and \textit{adaptive}. In Plot \ref{plt:appendix:isolated_agent}, we show the accuracy of the judge in predicting the preferred action when evidence was proposed by an isolated agent trained with one of these approaches for $1000$ different debate games. Without the confuser, both approaches amplify the capabilities of the judge, performing roughly equivalent. When faced with a confuser agent, the precommit approach performs better, since in the adaptive case, the agent is evaluated in a debate-like setup, which was not accounted for during training. In both cases, however, we observe that the isolated agent is not robust to an adversary.

\subsection{Alternative Utility Functions}\label{appendix:additional_results:rewdiff}
In Sec. \ref{sec:formal:debate}, we have defined the utility function $\mathbb{U}$ to output binary values $\{-1, 0, +1\}$ based on the justifiability rewards obtained from the judge. One might ask if there are alternative ways of defining $\mathbb{U}$ that preserve the zero-sum structure and potentially positively influence the learning. In this section, we consider one particular alternative that defines the utility function based on the difference between predicted rewards, which might provide a more informative learning signal. In particular, we set $u_1(n_L) = -u_2(n_L) = \mathbb{U}(a_t, a_t^B, \{e_{n_L}\})$, with $\mathbb{U}$ defined as $\mathbb{U}(a_t, a_t^B, \{e\}) = \mathcal{J}(a_t, \{e\}) - \mathcal{J}(a_t^B, \{e\})$. We rerun the experiments from Sec. \ref{sec:exp:argumentation} for maxmin and self-play agents and show preference recovery success and robustness results in Plot \ref{plt:appendix:argumentation_rewdiff}. We can see that the two approaches perform similarly in situations which do not involve a confuser agent. However, it seems that defining the utility using the difference in judge’s rewards leads to slightly lower scores when debate agents are faced with an adversary.

\subsection{Alternative Preference Dataset}\label{appendix:additional_results:dataset}
In Sec. \ref{sec:exp:environment:dataset} we construct a preference dataset by matching every pair $(s_t, a_t)$ from the cohort with an alternative action $a_r \sim \mathcal{U}(A), a_r \ne a_t$ sampled uniform-random from the action space. In this section, we examine additional strategies one could take when constructing such a synthetic dataset. In particular, we consider the \textit{random} strategy we just described, paired with two alternative ones, namely \textit{exhaustive} and \textit{offset}. The exhaustive strategy pairs $a_t$ with all possible alternative actions, $24$ in total. The offset strategy pairs $a_t$ with an alternative action that is in its neighborhood. To define a neighborhood, we recall that there are a total of $5$ choices for both, vasopressors (VC) and intravenous fluids (VC). Therefore, we can write $a_t = 5 * \text{IV} + \text{VC}$, where $\text{IV}, \text{VC} \in \{0, 1, 2, 3, 4\}$. To obtain an alternative action, we consider changing IV and VC by an offset sampled uniform-random from a set $\{-1, 0, 1\}$, for both IV and VC. We then train a new judge for each of the datasets and show its accuracy in Plot \ref{plt:appendix:dataset_generation_method}. The exhaustive variant represents the most informative, but also unrealistically large, dataset. The random variant represents somewhat of a “middle ground” in terms of dataset difficulty. Lastly, the offset variant represents the most difficult case, as differences between two actions are more nuanced. However, while the achieved accuracies reflect the difficulty of the corresponding dataset, the capabilities of a trained judge model are roughly the same.

\end{document}